\documentclass[journal]{IEEEtran}
\usepackage{cite}

\usepackage[colorlinks,linkcolor=red,anchorcolor=yellow,citecolor=green]{hyperref}
\usepackage{amssymb}
\usepackage{soul}
\usepackage{multirow}
\usepackage{multicol}
\usepackage{mathrsfs}
\usepackage{booktabs}
\usepackage{setspace}
\usepackage{color}

\ifCLASSINFOpdf
   \usepackage[pdftex]{graphicx}
   \graphicspath{{/figures/}}
   \DeclareGraphicsExtensions{.jpg,.eps}
\else
\fi
\usepackage{amsmath}
\DeclareMathOperator*{\argminA}{arg\,min}

\begin{document}
%
\title{Linking Points With Labels in 3D: A Review of Point Cloud Semantic Segmentation}
%
%
%

\author{Yuxing~Xie,
        Jiaojiao~Tian,~\IEEEmembership{Member,~IEEE}
        and~Xiao~Xiang~Zhu,~\IEEEmembership{Senior~Member,~IEEE}}

%
%

\markboth{IEEE Geoscience and Remote Sensing Magzine, Preprint.}%
{Shell \MakeLowercase{\textit{et al.}}: A Review of Point Cloud Semantic Segmentation}
%



\maketitle
\textbf{This is the preprint version in September 2019. To read the final version please go to IEEE Geoscience and
Remote Sensing Magazine on IEEE XPlore:}
\url{https://ieeexplore.ieee.org/document/9028090}\textbf{. If you would like to cite our work, please use the BibTeX as follows:}

\textcolor{blue}{@article\{xie2020linking,}

  \qquad \textcolor{blue}{title=\{Linking Points With Labels in 3D: A Review of Point Cloud Semantic Segmentation\},}
  
  \qquad \textcolor{blue}{author=\{Xie, Yuxing and Tian, Jiaojiao and Zhu, Xiao Xiang\},}
  
  \qquad \textcolor{blue}{journal=\{IEEE Geoscience and Remote Sensing Magazine\},}
  
  \qquad \textcolor{blue}{year=\{2020\},}
  
  \qquad \textcolor{blue}{publisher=\{IEEE\},}
  
  \qquad \textcolor{blue}{doi=\{10.1109/MGRS.2019.2937630\}}
  
\textcolor{blue}{\}}

\begin{abstract}
3D Point Cloud Semantic Segmentation (PCSS) is attracting increasing interest, due to its applicability in remote sensing, computer vision and robotics, and due to the new possibilities offered by deep learning techniques. In order to provide a needed up-to-date review of recent developments in PCSS, this article summarizes existing studies on this topic. Firstly, we outline the acquisition and evolution of the 3D point cloud from the perspective of remote sensing and computer vision, as well as the published benchmarks for PCSS studies. Then, traditional and advanced techniques used for Point Cloud Segmentation (PCS) and PCSS are reviewed and compared. Finally, important issues and open questions in PCSS studies are discussed. 
\end{abstract}


%
\IEEEpeerreviewmaketitle

\section{Motivation}
%
%
%
%
Semantic segmentation, in which pixels are associated with semantic labels, is a fundamental research challenge in image processing. Point Cloud Semantic Segmentation (PCSS) is the 3D form of semantic segmentation, in which regular or irregular distributed points in 3D space are used instead of regular distributed pixels in a 2D image. The point cloud can be acquired directly from sensors with distance measurability, or generated from stereo- or multi-view imagery. Due to recently developed stereovision algorithms and the deployment of all kinds of 3D sensors, point clouds, basic 3D data, have become easily accessible. High-quality point clouds provide a way to connect the virtual world to the real one. Specifically, they generate 2.5D/3D geometric structures, with which modeling is possible.


\subsection{Segmentation, classification, and semantic segmentation}
Research on PCSS has a long tradition involving different fields and defining distinct concepts for similar tasks. A brief clarification of some concepts is therefore necessary to avoid misunderstandings. The term PCSS is widely used in computer vision, especially in recent deep learning applications \cite{qi2017pointnet,landrieu2018large, wang2018dynamic}. However, in photogrammetry and remote sensing, PCSS is usually called ``point cloud classification" \cite{zhang2013svm,weinmann2015contextual,wang2015multiscale}. Or in some cases, this task is also called ``point labeling" \cite{koppula2011semantic,lu2012simplified,boulch2017unstructured}. In this article, to avoid confusion and to make this literature review keep up with latest deep learning techniques, we refer to point cloud semantic segmentation/classification/labeling, i.e., the task of associating each point of a point cloud with a semantic label, as PCSS.

Before effective supervised learning methods were widely applied in semantic segmentation, unsupervised Point Cloud Segmentation (PCS) was a significant task for 2.5D/3D data. PCS aims at grouping points with similar geometric/spectral characteristics without considering semantic information. In the PCSS workflow, PCS can be utilized as a presegmentation step, influencing the final results. Hence, PCS approaches are also included in this paper.

Single objects or the same classes of structures cannot be acquired from a raw point cloud directly. However, instance-level or class-level objects are required for object recognition. For example, urban planning and Building Information Modeling (BIM) need buildings and other man-made ground objects for reference \cite{tang2010automatic,volk2014building}. Forest remote sensing monitoring needs individual tree information based on their geometric structures \cite{lim2003lidar,wallace2012development}. Robotics applications, like Simultaneous Localization And Mapping (SLAM), need detailed indoor objects for mapping \cite{koppula2011semantic,rusu2008towards}. In some applications related to computer vision, such as autonomous driving, object detection, segmentation, and classification are necessary with the construction of a High Definition (HD) Map \cite{chen2017multi}. For the mentioned cases, PCSS and PCS are basic and critical tasks for 3D applications.

\subsection{New challenges and possibilities}
Papers\cite{nguyen20133d} and \cite{grilli2017review} provide two of the best available reviews for PCS and PCSS, but lack detailed information, especially for PCSS. Futhermore, in the past two years, deep learning has largely driven studies in PCSS. To meet the demand of deep learning, 3D datasets have improved, both in quality and diversity. Therefore, an updated study on current PCSS techniques is necessary.
This paper starts with the introduction of existing techniques to acquire point clouds and the existing benchmarks for point cloud study (section \ref{An Introduction to Point Cloud}). In section \ref{point cloud segmentation techniques} and \ref{point cloud semantic segmentation}, the major categories of algorithms are reviewed, for both PCS and PCSS. In section \ref{Discussion}, some issues related to data and techniques are discussed. Section \ref{Conclusion} concludes this paper with a technical outlook.

\section{An Introduction to Point Cloud} \label{An Introduction to Point Cloud}
\subsection{Point cloud data acquisition}
In computer vision and remote sensing, point clouds can be acquired with four main techniques: 1) Image-derived methods; 2) Light Detection And Ranging (LiDAR) systems; 3) Red Green Blue -Depth (RGB-D) cameras; and 4) Synthetic Aperture Radar (SAR) systems. Due to the differences in survey principles and platforms, their data features and application ranges are very diverse. A brief introduction to these techniques is provided below.


\subsubsection{Image-derived point cloud} 
Image-derived methods generate a point cloud indirectly from spectral imagery. First, they acquire stereo- or multi-view images through electro-optical systems, e.g., cameras. Then they calculate 3D isolated point information according to principles in photogrammetry or computer vision theory, either automatically or semi-automatically \cite{baltsavias1999comparison,westoby2012structure}. Based on distinct platforms, stereo- and multi-view image-derived systems can be divided into airborne, spaceborne, UAV-based, and close-range categories.

Early aerial traditional photogrammetry produced 3D points with semi-automatic human-computer  interaction in digital photogrammetric systems, characterized by strict geometric constraints and high survey accuracy \cite{mikhail2001introduction}. To produce this type of point data was time expensive due to many manual works. Therefore it was not feasible to generate dense points for large areas in this way. In the surveying and remote sensing industry, those early-form ``point clouds" were used in mapping and producing Digital Surface Models (DSMs) and Digital Elevation Models (DEMs). Due to the limitation of image resolutioan and the ability of processing multi-view images, traditional photogrammetry could only acquire close to nadir views with few building fa\c{c}ades from aerial/satellite platforms, which only generated a 2.5D point cloud rather than full 3D. At this stage, photogrammetry principles could also be applied as close-range photogrammetry in order to obtain points from certain objects or small-area scenes, but manual editing would also be necessary in the point cloud generating procedure.

Dense matching \cite{hirschmuller2005accurate,hirschmuller2008stereo,hirschmuller2007evaluation}, Multiple View Stereovision (MVS) \cite{furukawa2010accurate,nex2014uav}, and Structure from Motion (SfM) \cite{snavely2006photo,snavely2008modeling,westoby2012structure}, changed the image-derived point cloud, and opened the era of multiple view stereovision. SfM can estimate camera positions and orientations automatically, making it capable of processing multiview images simultaneously, while dense matching and MVS algorithms provide the ability to generate large volume of point clouds. In recent years, city-scale full 3D dense point clouds can be acquired easily through an oblique photography technique based on SfM and MVS. However, the quality of point clouds from SfM and MVS is not as good as those generated by traditional photogrammetry or LiDAR techniques, and it is especially unreliable for large regions \cite{xiao2013sun3d}. 

Compared to airborne photogrammetry, satellite stereo system is disadvantaged in terms of spatial resolution and availability of multi-view imagery. However, satellite cameras are able to map large regions in a short period of time with relatively lower cost. Also due to new dense matching techniques and their improved spatial resolution, satellite imagery is becoming an important data source for image-derived point clouds.

\subsubsection{LiDAR point cloud}
Light Detection And Ranging (LiDAR) is a surveying and remote sensing technique. As its name suggests, LiDAR utilizes laser energy to measure the distance between the sensor and the object to be surveyed \cite{shan2018topographic}. Most LiDAR systems are pulse-based. The basic principle of pulse-based measuring is to emit a pulse of laser energy and then measure the time it takes for that energy to travel to a target. Depending on sensors and platforms, the point density or resolution varies greatly, from less than 10 points per $m^2$ ($pts/m^2$) to thousands of points per $m^2$ \cite{qin20163d}. Based on platforms, LiDAR systems are divided into airborne LiDAR scanning (ALS), terrestrial LiDAR scanning (TLS), mobile LiDAR scanning (MLS) and unmanned LiDAR scanning (ULS) systems.

ALS operates from airborne platforms. Early ALS LiDAR data are 2.5D point clouds, which are similar to traditional photogrammetric point clouds. The density of ALS points is normally low, as the distance from an airborne platform to the ground is large. In comparison to traditional photogrammetry, ALS point clouds are more expensive to acquire and normally contain no spectral information. Vaihingen point cloud semantic labeling dataset \cite{rottensteiner2013isprs} is a typical ALS benchmark dataset. Multispectral airborne LiDAR is a special form of an ALS system that obtains data using different wavelengths.  Multispectral LiDAR performs well for the extraction of water, vegetation and shadows, but the data are not easily available \cite{morsdorf2009assessing,wallace2012recovery}.

TLS, also called static LiDAR scanning, scans with a tripod-mounted stationary sensor. Since it is used in a middle- or close-range environment, the point cloud density is very high. Its advantage is its ability to provide real, high quality 3D models. Until now TLS has been commonly used for modeling small urban or forest sites, and heritage or artwork documentation. Semantic3D.net\cite{hackel2017semantic3d} is a typical TLS benchmark dataset. 

MLS operates from a moving vehicle on the ground, with the most common platforms being cars. Currently, research and development on autonomous driving is a hot topic, for which HD maps are essential. The generation of HD maps is therefore the most significant application for MLS. Several mainstream point cloud benchmark datasets belong to MLS \cite{bredif2014terramobilita,roynard2018paris}.

ULS systems are usually deployed on drones or other unmanned vehicles. Since they are relatively cheap and very flexible, this recent addition to the LiDAR family is currently becoming more and more popular. Compared to ALS, where the platform is working above the objects, ULS can provide a shorter-distance LiDAR survey application, collecting denser point clouds with higher accuracy. Thanks to the small size and light weight of its platform, ULS offers high operational flexibility. Therefore, in addition to traditional LiDAR tasks (e.g., acquiring DSMs), ULS has advantages in agriculture and forestry surveying, disaster monitoring and mining surveying \cite{sankey2017uav,zhang2019automated,li20193d}.

For LiDAR scanning, since the system is always moving with the platform, it is necessary to combine points' positions with Global Navigation Satellite System (GNSS) and Inertial Measurement Unit (IMU) data to ensure a high-quality matching point cloud. Until now, LiDAR has been the most important data source for point cloud research and has been used to provide ground truth to evaluate the quality of other point clouds. 

\subsubsection{RGB-D point cloud}
An RGB-D camera is a type of sensor that can acquire both RGB and depth information. There are three kinds of RGB-D sensors, based on different principles: (a) structured light \cite{han2013enhanced}, (b) stereo \cite{mattoccia2015passive}, and (c) time of flight \cite{lachat2015first}. Similar to LiDAR, the RGB-D camera can measure the distance between the camera to the objects, but pixel-wise. However, an RGB-D sensor is much cheaper than a LiDAR system. Microsoft's Kinect is a well-known and widely used RGB-D sensor \cite{han2013enhanced,lachat2015first}. In an RGB-D camera, relative orientation elements between or among different sensors are calibrated and known, so co-registered synchronized RGB images and depth maps can be easily acquired. Obviously, the point cloud is not the direct product of RGB-D scanning. But since the position of the camera's center point is known, the 3D space position of each pixel in a depth map can be easily obtained, and then directly used to generate the point cloud. RGB-D cameras have three main applications: object tracking, human pose or signature recognition, and SLAM-based environment reconstruction. Since mainstream RGB-D sensors are close-range, even much closer than TLS, they are usually employed in indoor environments. Several mainstream indoor point cloud segmentation benchmarks are RGB-D data \cite{dai2017scannet,armeni20163d}.

\subsubsection{SAR point cloud}
Interferometric Synthetic Aperture Radar (InSAR), a radar technique crucial to remote sensing, generates maps of surface deformation or digital elevation based on the comparison of multiple SAR image pairs. A rising star, InSAR-based point cloud has showed its value over the past few years and is creating new possibilities for point cloud applications \cite{shahzad2012facade,zhu2014facade,shahzad2015robust,shahzad2015segmentation,schmitt2015reconstruction}. Synthetic Aperture Radar tomography (TomoSAR) and Persistent Scatterer Interferometry (PSI) are two major techniques that generate point clouds with InSAR, extending the principle of SAR into 3D \cite{bamler2009interferometric,zhu2010very}. Compared with PSI, TomoSAR's advantage is its detailed reconstruction and monitoring of urban areas, especially man-made infrastructure \cite{zhu2010very}. The TomoSAR point cloud has a point density that is comparable to ALS LiDAR \cite{gernhardt2010potential,gernhardt2012geometrical}. These point clouds can be employed for applications in building reconstruction in urban areas, as they have the following features \cite{zhu2014facade}:

(a) TomoSAR point clouds reconstructed from spaceborne data have a moderate 3D positioning accuracy on the order of 1 m \cite{zhu2012super}, even able to reach a decimeter level by geocoding error correction techniques\cite{montazeri2018geocoding}, while ALS LiDAR provides accuracy typically on the order of 0.1 m \cite{rottensteiner2002new}. 

(b) Due to their coherent imaging nature and side-looking geometry, TomoSAR point clouds emphasize different objects with respect to LiDAR systems: a) The side-looking SAR geometry enables TomoSAR point clouds to possess rich fa\c{c}ade information: results using pixel-wise TomoSAR for the high-resolution reconstruction of a building complex with a very high level of detail from spaceborne SAR data are presented in \cite{zhu2012demonstration}; b) temporarily incoherent objects, e.g., trees, cannot be reconstructed from multipass spaceborne SAR image stacks; and c) to obtain the full structure of individual buildings from space, fa\c{c}ade reconstruction using TomoSAR point clouds from multiple viewing angles is required \cite{shahzad2012facade, zhu2012tomosar}.

(c) Complementary to LiDAR and optical sensors, SAR is so far the only sensor capable of providing fourth dimension information from space, i.e., temporal deformation of the building complex \cite{zhu2011let}, and microwave scattering properties of the fa\c{c}ade reflect geometrical and material features.

InSAR point clouds have two main shortcomings that affect their accuracy: (1) Due to limited orbit spread and the small number of images, the location error of TomoSAR points is highly anisotropic, with an elevation error typically one or two orders of magnitude higher than in range and azimuth; (2) Due to multiple scattering, ghost scatterers may be generated, appearing as outliers far away from a realistic 3D position \cite{auer2011ghost}.

Compared with the aforementioned image-derived, LiDAR-based, and RGB-D-based point cloud, the data from SAR have not yet been widely used for studies and applications. However, mature SAR satellites, such as TerraSAR-X, have collected rich global SAR data, which are available for InSAR-based reconstruction at global scale\cite{shi2019nonlocal}. Hence, the SAR point cloud can be expected to play a conspicuous role in the future.

\subsection{Point cloud characters}
From the perspective of sensor development and various applications, we have cataloged point clouds into: (a) sparse (less than 20 $pts/m^2$), (b) dense (hundreds of $pts/m^2$), and (c) multi-source.

(a) In their early stage, which was limited by matching techniques and computation ability, photogrammetric point clouds were sparse and small in volume. At that time, laser scanning systems had limited types and were not widely used. ALS point clouds, mainstream laser data, were also sparse. Limited by the point density, point clouds at this stage were not able to represent land covers in object level. Therefore there was no specific demand for precise PCS or PCSS. Researchers mainly focused on 3D mapping (DEM generation), and simple object extraction (e.g., rooftops).

(b) Computer vision algorithms, such as dense matching, and high-efficiency point cloud generators, such as various LiDAR systems and RGB-D sensors, opened the big data era of the dense point cloud. Dense and large-volume point clouds created more possibilities in 3D applications but also had a stronger desire for practicable algorithms. PCS and PCSS were newly proposed and became increasingly necessary, since only a class-level or instance-level point cloud further connect virtual word to the real one. Both computer vision and remote sensing need PCS and PCSS solutions to develop class-level interactive applications. 

(c) From the perspective of general computer vision, research on the point cloud and its related algorithms remain at stage (b). However, as a benefit to the development of spaceborne platforms and multi-sensors, remote sensing researchers developed a new understanding of the point cloud. New-generation point clouds, such as satellite photogrammetric point clouds and TomoSAR point clouds, stimulated demand for relevant algorithms. Multi-source data fusion has become a trend in remote sensing \cite{wang2014automatic,schmitt2016data,wang2017fusing}, but current algorithms in computer vision are insufficient for such remote sensing datasets. To fully exploit multi-source point cloud data, more research is needed.

As we have reviewed, different point clouds have different features and application environments. Table \ref{tab:table1} provides an overview of basic information about various point clouds, including point density, advantages, disadvantages, and applications. 

\begin{table*}[!phtb]
 \caption{An overview of various point clouds}
  \centering\onehalfspacing\small
  \begin{tabular}{|p{30pt}|p{30pt}|p{100pt}|p{100pt}|p{100pt}|p{100pt}|}\hline
 \multicolumn{2}{|c|}{} &  \begin{center}\textbf{Point density}\end{center} & \begin{center}\textbf{Advantages}\end{center} & \begin{center}\textbf{Disadvantages}\end{center} & \begin{center}\textbf{Applications}\end{center}  \\ \hline
  
  \multicolumn{2}{|c|}{\textbf{Image-derived}} & From sparse ($\textless10 pts/m^2$) to very high ($\textgreater400 pts/m^2$), depending on the spatial resolution of  the stereo- or multi-view images & With color (RGB, multi-spectral) information; suitable for large area (airborne, spaceborne) & Influenced by light; accuracy depends on available precise camera models, image matching algorithms, stereo angles, image resolution and image quality; not suitable for areas or objects without texture, such as water or snow-covered regions; influenced by shadows in images & Urban monitoring; vegetation monitoring; 3D object reconstruction; etc. \\ \hline

    & \begin{center}\textbf{ALS}\end{center} & \begin{center}Sparse ($\textless20 pts/m^2$); when the survey distance is shorter, the density is higher\end{center} & High accuracy ($\textless15cm$); suitable for large area; not affected by weather &  & Urban monitoring; vegetation monitoring; power line detection; etc. \\ 
   \cline{2-4}\cline{6-6}   
   \textbf{LiDAR} & \begin{center}\textbf{MLS}\end{center} & Dense ($\textgreater100 pts/m^2$); when the survey distance is shorter, the density is higher & High accuracy (cm-level) & Expensive; affected by mirror reflection; long scanning time & HD map; urban monitoring  \\
   \cline{2-4}\cline{6-6}   
    & \begin{center}\textbf{TLS}\end{center} & Dense ($\textgreater100 pts/m^2$); when the survey distance is shorter, the density is higher & High accuracy (mm-level) & & Small-area 3D reconstruction \\
     \cline{2-4}\cline{6-6} 
       & \begin{center}\textbf{ULS}\end{center} & Dense ($\textgreater100 pts/m^2$); when the survey distance is shorter, the density is higher & High accuracy (cm-level) & & Forestry survey; mining survey; disaster monitoring; etc.   \\
    \hline
    
    \multicolumn{2}{|c|}{\textbf{RGB-D}} & \center Middle-density & Cheap; flexible & Close-range; limited accuracy & Indoor reconstruction; object tracking; human pose recognition; etc. \\
    
    \hline
    
    \multicolumn{2}{|c|}{\textbf{InSAR}} & \center Sparse ($\textless20 pts/m^2$) & Global data is available; compared to ALS, complete building fa\c{c}ade information is available; 4D information; middle-accuracy;  not affected by weather & Expensive data; ghost scatterers; preprocessing techniques are needed & Urban monitoring; forest monitoring; etc. \\
    
    \hline

  \end{tabular}
  \label{tab:table1}
\end{table*}

\subsection{Point cloud application}
In the studies on PCS and PCSS, data and algorithm selections are driven by the requirements of specific applications. In this section, we outline most of the studies focusing on PCS and PCSS reviewed in this article (see Table \ref{tab:table2}). These works are classified according to their point cloud data types and working environments. The latter include urban, forest, industry, and indoor settings. In Table \ref{tab:table2}, texts in brackets, after each reference, contain the corresponding publishing year and main methods. Algorithm types are represented as abbreviations.

\begin{table*}[!phtb]
 \caption{An overview of PCS and PCSS applications sorted according to data acquisitions}
  \onehalfspacing\footnotesize
    RG is short for Region Growing. HT is short for Hough Transform. R is short for RANSAC. C is short for Clustering-based. O is short for Oversegnentation. ML is short for Machine Learning. DL is short for Deep Learning.
 \begin{center}
  \begin{tabular}{p{60pt}p{200pt}p{60pt}p{60pt}p{110pt}}     \hline
                   &        \qquad \qquad \qquad \qquad \qquad \textbf{Urban}       &   \qquad \textbf{Forest}     &    \qquad  \textbf{Industry}       &      \qquad \qquad \textbf{Indoor}   \\
   \hline               
   \begin{center} \textbf{Image-derived}\end{center}   &  \textbf{Building fa\c{c}ades}:\cite{adam2018h} (2018/RG), \cite{bauer2005segmentation} (2005/RG);  
   \textbf{PCSS}: \cite{boulch2018snapnet} (2018/DL), \cite{su2018splatnet} (2018/DL), \cite{riegler2017octnet} (2017/DL), \cite{choy20194d} (2019/DL) &   &   & \textbf{Plane PCS}: \cite{limberger2015real} (2015/HT) \\
   \hline
   \begin{center}\textbf{ALS}\end{center}  &  \textbf{Building plane PCS}: \cite{xu2015investigation} (2015/R), \cite{chen2014methodology} (2014/R), \cite{tarsha2007hough} (2007/R, HT), 
   \cite{gorte2002segmentation} (2002/HT), \cite{sampath2006clustering} (2006/C), \cite{sampath2010segmentation} (2010/C), 
   \cite{ural2012min} (2012/C), \cite{yan2014global} (2014/C);  \textbf{Urban scene}: \cite{melzer2007non} (2007/C), \cite{yao2009object} (2009/C); 
    \textbf{PCSS}: \cite{lodha2007aerial} (2007/ML), \cite{carlberg2009classifying} (2009/ML), \cite{chehata2009airborne} (2009/ML), \cite{shapovalov2010nonassociative} (2010/ML),
   \cite{niemeyer2012conditional} (2012/ML), \cite{niemeyer2014contextual} (2014/ML), \cite{vosselman2017contextual} (2017/HT, R, ML), \cite{xiong20113} (2011/ML),
   \cite{najafi2014non} (2014/ML), \cite{zhang2013svm} (2013/HT, ML)
   & \textbf{Tree structure PCS}: \cite{morsdorf2004lidar}(2004/C);  \textbf{Forest structure}: \cite{ferraz20103d} (2010/C)
   &   &  \\
   \hline
   \begin{center}\textbf{MLS}\end{center}  &  \textbf{Buildings}: \cite{vo2015octree} (2015/RG); \textbf{Urban objects}: \cite{nurunnabi2012robust} (2012/RG); \textbf{PCSS}: \cite{xiong20113} (2011/ML), \cite{weinmann2015semantic} (2015/ML), \cite{weinmann2015contextual} (2015/ML), \cite{lu2012simplified} (2012/ML), \cite{najafi2014non} (2014/ML), \cite{munoz2009contextual} (2009/ML), \cite{landrieu2017structured} (2017/ML), \cite{tchapmi2017segcloud} (2017/DL), \cite{ye20183d} (2018/DL), \cite{landrieu2019point} (2019/O, DL) 
   & & & \textbf{Plane PCS}: \cite{xiao2013three} (2013/R), \cite{li2017improved} (2017/R) \\
   \hline
   \begin{center}\textbf{TLS}\end{center}  &  \textbf{Building/building structure PCS}: \cite{boulaassal2007automatic} (2007/R), \cite{vo2015octree} (2015/RG), \cite{dong2018efficient} (2018/RG, C), \cite{biosca2008unsupervised} (2008/C); \textbf{Buildings and trees}: \cite{ning2009segmentation} (2009/RG); \textbf{Urban scene}: \cite{xu2016segmentation} (2016/O, C), \cite{xu2017voxel} (2017/O, C), \cite{xu2018unsupervised} (2018/O, C); \textbf{PCSS}: \cite{wang2015multiscale} (2015/ML), \cite{lim20093d} (2009/O, ML), \cite{li2016three} (2016/ML), \cite{boulch2018snapnet} (2018/DL), \cite{tchapmi2017segcloud} (2017/DL), \cite{landrieu2018large} (2018/O, DL), \cite{wang2019graph} (2019/DL)\cite{choy20194d} (2019/DL)
   
   & \textbf{Tree PCSS}: \cite{lalonde2005scale} (2005/ML)
   &   &  \textbf{Plane PCS}:\cite{borrmann20113d} (2011/HT) \\
   \hline
   \begin{center}\textbf{RGB-D}\end{center}  &  &  &  &  \textbf{Plane PCS}: \cite{hulik2014continuous} (2014/HT), \cite{dong2018efficient} (2018/RG, C); \textbf{PCSS}: \cite{khoshelham2012accuracy} (2012/ML), \cite{shapovalov2013spatial} (2013/ML), \cite{huang2018recurrent} (2018/DL), \cite{li2018pointcnn} (2018/DL), \cite{tchapmi2017segcloud} (2017/DL), \cite{qi2017pointnet} (2017/DL), \cite{qi2017pointnet++} (2017/DL), \cite{wang2018dynamic} (2018/DL), \cite{landrieu2018large} (2018/DL), \cite{ye20183d} (2018/DL), \cite{jiang2018pointsift} (2018/DL), \cite{choy20194d} (2019/DL), \cite{wang2019graph} (2019/DL), \cite{Wu2019PointConv} (2019/DL), \cite{Wang2019Associatively} (2019/DL), \cite{pham2019jsis3d} (2019/DL), \cite{komarichev2019cnn} (2019/DL), \cite{zhao2019pointweb} (2019/DL), \cite{landrieu2019point} (2019/O, DL);
   \textbf{Instance segmentation}: \cite{wang2018sgpn} (2018/DL), \cite{yi2019gspn} (2019/DL), \cite{Wang2019Associatively} (2019/DL), \cite{pham2019jsis3d} (2019/DL) \\
   \hline
   \center \textbf{InSAR}  &  \textbf{Building/building structure}: \cite{shahzad2015robust} (2015/C), \cite{shahzad2012facade} (2012/C), \cite{zhu2014facade} (2014/C)  &  \textbf{Tree PCS}: \cite{shahzad2015segmentation} (2015/C)  & & \\
   \hline
   \center \textbf{Not mentioned data} & & &\cite{rabbani2005efficient}(2005/HT), \cite{tran2015extraction} (2015/R), \cite{le2018acquiring} (2018/R) & \\
   \hline
   \end{tabular}
   \end{center}
   \label{tab:table2}
\begin{center}
\end{center}
\end{table*}

Several issues can be summarized from Table \ref{tab:table2}:
(a) LiDAR point clouds are the most commomly used data in PCS. They have been widely used for buildings (urban environments) and trees (forests). Buildings are also the most popular research objects in traditional PCS. As buildings are usually constructed with regular planes, plane segmentation is a fundamental topic in building segmentation.

(b) Image-derived point clouds have been frequently used in real-world scenarios. However, mainly due to the limitation of available annotated benchmarks, there are not many PCS and PCSS studies on image-based data. Currently, there is only one public influential dataset based on image-derived points, whose range is only a very small area around one single building \cite{riemenschneider2014learning}. More efforts are therefore needed in this area.

(c) RGB-D sensors are limited by their close range, so they are usually applied in an indoor environment. In PCS studies, plane segmentation is the main task for RGB-D data. In PCSS studies, since there are several benchmark datasets from RGB-D sensors, many deep learning-based methods are tested on them.
 
(d) As for InSAR point clouds, although there are not many PCS or PCSS studies, these have shown potential in urban monitoring, especially building structure segmentation.

\subsection{Benchmark datasets} \label{Benchmark datasets}
Public standard benchmark datasets achieve significant effectiveness for algorithm development, evaluation and comparison. It should be noted that most of them are labeled for PCSS rather than PCS. Since 2009, several benchmark datasets have been available for PCSS. However, early datasets have plenty of shortcomings. For example, the Oakland outdoor MLS dataset \cite{munoz2009contextual}, the Sydney Urban Objects MLS dataset \cite{de2013unsupervised}, the Paris-rue-Madame MLS dataset \cite{serna2014paris}, the IQmulus \& TerraMobilita Contest MLS dataset \cite{bredif2014terramobilita} and ETHZ CVL RueMonge 2014 multiview stereo dataset\cite{riemenschneider2014learning} can not sufficiently provide both different object representations and labeled points. KITTI \cite{geiger2013vision} and NYUv2 \cite{silberman2012indoor} have more objects and points than the aforementioned datasets, but they do not provide a labeled point cloud directly. These must be generated from 3D bounding boxes in KITTI or depth images in NYUv2.

To overcome the drawbacks of early datasets, new benchmark data have been made available in recent years. Currently, mainstream PCSS benchmark datasets are from either LiDAR or RGB-D sensors. A nonexhaustive list of these datasets follows.

\subsubsection{Semantic3D.net}
The semantic3D.net \cite{hackel2017semantic3d} is a representative large-scale outdoor TLS PCSS dataset. It is a collection of urban scenes with over four billion labeled 3D points in total for PCSS purposes only. Those scenes contain a range of diverse urban objects, divided into eight classes, including man-made terrain, natural terrain, high vegetation, low vegetation, buildings, hardscape, scanning artefacts, and cars. In consideration of the efficiency of different algorithms, two types of sub-datasets were designed, semantic-8 and reduced-8. Semantic-8 is the full dataset, while reduced-8 uses training data in the same way as semantic-8, but only includes four small-sized subsets as test data. This dataset can be downloaded at \url{http://www.semantic3d.net/}. To learn the performance of different algorithms on this dataset, readers are recommended to refer to \cite{landrieu2018large,boulch2018snapnet,wang2019graph}.


\subsubsection{Stanford Large-scale 3D Indoor Spaces Dataset (S3DIS)}
Unlike semantic3D.net, S3DIS \cite{armeni20163d} is a large-scale indoor RGB-D dataset, which is also a part of the 2D-3D-S dataset \cite{armeni2017joint}. It is a collection of over 215 million points, covering an area of over 6,000 $m^2$ in six indoor regions originating from three buildings. The main covered areas are for educational and office use. Annotations in S3DIS have been prepared at an instance level. Objects are divided into structural and movable elements, which are further classified into 13 classes (structural elements: ceiling, floor, wall, beam, column, window, door; movable elements: table, chair, sofa, bookcase, board, clutter for all other elements). The dataset can be requested from \url{http://buildingparser.stanford.edu/dataset.html}. To learn the performance of different algorithms on this dataset, readers are recommended to refer to \cite{li2018pointcnn,landrieu2018large,choy20194d,landrieu2019point}.

\subsubsection{Vaihingen point cloud semantic labeling dataset}
This dataset \cite{rottensteiner2013isprs} is the most well-known published benckmark dataset in the remote sensing field in recent years. It is a collection of ALS point cloud, consisting of 10 strips captured by a Leica ALS50 system with a $45^{\circ}$ field of view and 500 $m$ mean flying height over Vaihingen, Germany.  The average overlap between two neighboring strips is around 30\% and the median point density is 6.7 $points/m^2$ \cite{rottensteiner2013isprs}. This dataset had no label at a point level at first. Niemeyer et al. \cite{niemeyer2014contextual} first used it for a PCSS test and labeled points in three areas. Now the labeled point cloud is divided into nine classes as an algorithm evaluation standard. Although this dataset has significantly fewer points compared with semantic3D.net and S3DIS, it is an influential ALS dataset for remote sensing. The dataset can be requested from \url{http://www2.isprs.org/commissions/comm3/wg4/3d-semantic-labeling.html}.

\subsubsection{Paris-Lille-3D}
The Paris-Lille-3D \cite{roynard2018paris} is a brand new benchmark for PCSS, as it was published in 2018. It is an MLS point cloud dataset with more than 140 million labelled points, including 50 different urban object classes along 2 $km$ of streets in two French cities, Paris and Lille. As an MLS dataset, it also could be used for autonomous vehicles. As this is a recent dataset, only a few validated results are shown on the related website. This dataset is available at \url{http://npm3d.fr/paris-lille-3d}.


\subsubsection{ScanNet}
ScanNet \cite{dai2017scannet} is an instance-level indoor RGB-D dataset that includes both 2D and 3D data. In contrast to the benchmarks mentioned above, ScanNet is a collection of labeled voxels rather than points or objects. Up to now, ScanNet v2, the newest version of ScanNet, has collected 1513 annotated scans with an approximate 90\% surface coverage. In the semantic segmentation task, this dataset is marked in 20 classes of annotated 3D voxelized objects. Each class corresponds to one category of furniture. This dataset can be requested from \url{http://www.scan-net.org/index\#code-and-data}. To learn the performance of different algorithms on this dataset, readers are recommended to refer to \cite{qi2017pointnet++,Wang2019Associatively,choy20194d,pham2019jsis3d}.


\section{point cloud segmentation techniques} \label{point cloud segmentation techniques}
PCS algorithms are mainly based on strict hand-crafted features from geometric constraints and statistical rules. The main process of PCS aims at grouping raw 3D points into non overlapping regions. Those regions correspond to specific structures or objects in one scene. Since no supervised prior knowledge is required in such a segmentation procedure, the delivered results have no strong semantic information. Those approaches could be categorized into four major groups:  edge-based, region growing, model fitting, and clustering-based.

\subsection{Edge-based}
Edge-based PCS approaches were directly transferred from 2D images to 3D point clouds, which were mainly used in the very early stage of PCS. As the shapes of objects are described by edges, PCS can be solved by finding the points that are close to the edge regions.  The principle of edge-based methods is to locate the points that have a rapid change in intensity \cite{nguyen20133d}, which is similar to some 2D image segmentation approaches.

According to the definition from \cite{rabbani2006segmentation}, an edge-based segmentation algorithm is formed by two main stages: (1) edge detection, where the boundaries of different regions are extracted, and (2) grouping points, where the final segments are generated by grouping points inside the boundaries from (1). For example, in \cite{bhanu1986range}, the authors designed a gradient-based algorithm for edge detection, fitting 3D lines to a set of points and detecting changes in the direction of unit normal vectors on the surface. In \cite{jiang1996fast}, the authors proposed a fast segmentation approach based on high-level segmentation primitives (curve segments), in which the amount
of data could be significantly reduced. Compared to the method presented in \cite{bhanu1986range}, this algorithm is both accurate and efficient, but it is only suitable for range images, and may not work for uneven-density point clouds. Moreover, paper \cite{sappa2001fast} extracted close contours from a binary edge map for fast segmentation. Paper \cite{wani2003parallel} introduced a parallel edge-based segmentation algorithm extracting three types of edges. An algorithm optimization mechanism, named reconfigurable multiRing network, was applied in this algorithm to reduce its runtime.

The edge-based algorithms enable a fast PCS due to its simplicity, but their good performance can only be maintained when simple scenes with ideal points are provided (e.g., low noise, even density). Some of them are only suitable for range images rather than 3D points. Thus this approach is rarely applied for dense and/or large-area point cloud datasets nowadays. Besides, in 3D space, such methods often deliver disconnected edges, which cannot be used to identify closed segments directly, without a filling or interpretation procedure \cite{castillo2013point,grilli2017review}.

\subsection{Region growing}
Region growing is a classical PCS method, which is still widely used today. It uses criteria, combining features between two points or two region units in order to measure the similarities among pixels (2D), points (3D), or voxels (3D), and merge them together if they are spatially close and have similar surface properties. Besl and Jain \cite{besl1988segmentation} introduced a two-step initial algorithm: (1) coarse segmentation, in which seed pixels are selected based on the mean and Gaussian curvature of each point and its sign; and (2) region growing, in which interactive region growing is used to refine the result of step (1) based on a variable order bivariate surface fitting. Initially, this method was primarily used in 2D segmentation. As in the early stage of PCS research most point clouds were actually 2.5D airborne LiDAR data, in which only one layer has a view in the $z$ direction, the general preprocessing step was to transform points from 3D space into a 2D raster domain \cite{geibel2000segmentation}. With the more easily available real 3D point clouds, region growing was soon adopted directly in 3D space. This 3D region growing technique has been widely applied in the segmentation of building plane structures \cite{gorte2002segmentation,nurunnabi2012robust,vo2015octree,xiao2013three,dong2018efficient}.

Similar to the 2D case, 3D region growing comprises two steps: (1) select seed points or seed units; and (2) region growing, driven by certain principles. To design a region growing algorithm, three crucial factors should be taken into consideration: criteria (similarity measures), growth unit, and seed point selection. For the criteria factor, geometric features, e.g., Euclidean distance or normal vectors, are commonly used. For example, Ning et al. \cite{ning2009segmentation} employed the normal vector as criterion, so that the coplanar may share the same normal orientation. Tovari et al. \cite{tovari2005segmentation} applied normal vectors, the distance of the neighboring points to the adjusting plane, and the distance between the current point and candidate points as the criteria for merging a point to a seed region that was randomly picked from the dataset after manually filtering areas near edges. Dong et al. \cite{dong2018efficient} chose normal vectors and the distance between two units.

 For growth unit factor, there are usually three strategies: (1) single points, (2) region units, e.g., voxel grids and octree structures, and (3) hybrid units. Selecting single points as region units was the main approach in the early stages \cite{rabbani2006segmentation,ning2009segmentation}. However, for massive point clouds, point-wise calculation is time-consuming. To reduce the data volume of the raw point cloud and improve calculation efficiency, e.g., neighborhood search with a $k$-d tree in raw data \cite{deschaud2010fast}, the region unit is an alternative idea of direct points in 3D region growing. In a point cloud scene, the number of voxelized units is smaller than the number of points. In this way, the region growing process can be accelerated significantly. Guided by this strategy, Deschaud et al. \cite{deschaud2010fast} presented a voxel-based region growing algorithm to improve efficiency by replacing points with voxels during the region growing procedure. Vo et al. \cite{vo2015octree} proposed an adaptive octree-based region growing algorithm for fast surface patch segmentation by incrementally grouping adjacent voxels with a similar saliency feature. As a balance of accuracy and efficiency, hybrid units were also proposed and tested by several studies. For example, Xiao et al. \cite{xiao2013three} combined single points with subwindows as growth units to detect planes. Dong et al. \cite{dong2018efficient} utilized a hybrid region growing algorithm, based on units of both single points and supervoxels, to realize coarse segmentation before global energy optimization. 

For Seed point selection, since many region growing algorithms aim at plane segmentation, a usual practice is designing a fitting plane for a certain point and its neighbor points first, and then choosing the point with minimum residual to the fitting plane as a seed point \cite{rabbani2006segmentation,ning2009segmentation}. The residual is usually estimated by the distance between one point and its fitting plane \cite{rabbani2006segmentation,ning2009segmentation} or the curvature of the point \cite{nurunnabi2012robust,dong2018efficient}. 

Nonuniversality is a nontrivial problem for region growing \cite{vo2015octree}. The accuracy of these algorithms relies on the growth criteria and locations of the seeds, which should be predefined and adjusted for different datasets. In addition, these algorithms are computationally intensive and may require a reduction in data volume for a trade-off between accuracy and efficiency.

\subsection{Model fitting}
The core idea of model fitting is matching the point clouds to different primitive geometric shapes, thus it has been normally regarded as a shape detection or extraction method. However, when dealing with scenes with parameter geometric shapes/models, e.g., planes, spheres, and cylinders, model fitting can also be regarded as a segmentation approach. Most widely used model-fitting methods are built on two classical algorithms, Hough Transform (HT) and RANdom SAmple Consensus (RANSAC).

\subsubsection{HT}
HT is a classical feature detection technique in digital image processing. It was initially presented in \cite{hough1962method} for line detection in 2D images. There are three main steps in HT \cite{xu1990new}: (1) mapping every sample (e.g., pixels in 2D images and points in point clouds) of the original space into a discretized parameter space; (2) laying an accumulator with a cell array on the parameter space and then, for each input sample, casting a vote for the basic geometric element of which they are inliers in the parameter space; and (3) selecting the cell with the local maximal score, of which parameter coordinates are used to represent a geometric segment in original space. The most basic version of HT is Generalized Hough Transform (GHT), also called the Standard Hough Transform (SHT), which is introduced in \cite{duda1972use}. GHT uses an angle-radius parameterization instead of the original slope-intercept form, in order to avoid the infinite slope problem and simplify the computation. The GHT is based on:
\begin{equation}\rho=x\cos(\theta)+y\sin(\theta)\end{equation}
where $x$ and $y$ are the image coordinates of a corresponding sample pixel, $\rho$ is the distance between the origin and the line through the corresponding pixel, and $\theta$ is the angle between the normal of the above-mentioned line and the $x$-axis. Angle-radius parameterization can also be extended into 3D space, and thus can be used in 3D feature detection and regular geometric structure segmentation. Compared with the 2D form, in 3D space, there is one more angle parameter, $\phi$:
\begin{equation}\rho=x\cos(\theta)\sin(\phi)+y\sin(\theta)\sin(\phi)+z\cos(\phi)\end{equation}
where $x$, $y$, and $z$ are corresponding coordinates of a 3D sample (e.g., one specific point from the whole point cloud), and $\theta$ and $\phi$ are polar coordinates of the normal vector of the plane, which includes the 3D sample.

One of the major disadvantages of GHT is the lack of boundaries in the parameter space, which leads to high memory consumption and long calculation time \cite{kaiser2018survey}. Therefore, some studies have been conducted to improve the performance of HT by reducing the cost of the voting process \cite{limberger2015real}. Such algorithms include Probabilistic Hough transform (PHT) \cite{kiryati1991probabilistic}, Adaptive probabilistic Hough transform (APHT) \cite{yla1994adaptive}, Progressive Probabilistic Hough Transform (PPHT) \cite{galamhos1999progressive}, Randomized Hough Transform (RHT) \cite{xu1990new}, and Kernel-based Hough Transform (KHT) \cite{fernandes2008real}. In addition to computational costs, choosing a proper accumulator representation is also a way to optimize HT performance \cite{borrmann20113d}.

Several review articles involving 3D HT are available \cite{borrmann20113d,kaiser2018survey,limberger2015real}. As with region growing in the 3D field, planes are the most frequent research objects in HT-based segmentation \cite{vosselman2004recognising,tarsha2007hough,hulik2014continuous,limberger2015real}. In addition to planes, other basic geometric primitives can also be segmented by HT. For example, Rabbani et al. \cite{rabbani2005efficient} used a Hough-based method to detect cylinders in point clouds, similar to plane detection. In addition, a comprehensive introduction to sphere recognition based on HT methods is presented in \cite{camurri20143d}. 

To evaluate different HT algorithms on point clouds, Borrmann et al. \cite{borrmann20113d} compared improved HT algorithms and concluded that RHT was the best one for PCS at that time, due to its high efficiency. Limberger et al. \cite{limberger2015real} extended KHT \cite{fernandes2008real} to 3D space, and proved that 3D KHT performed better than previous HT techniques, including RHT, for plane detection. The 3D KHT approach is also robust to noise and even to irregularly distributed samples \cite{limberger2015real}.

\subsubsection{RANSAC}
The RANSAC technique is the other popular model fitting method\cite{fischler1981random}. Several reviews about general RANSAC-based methods have been published. Learning more about the RANSAC family and their performance is highly recommended, particularly in \cite{choi2009performance,raguram2008comparative,raguram2013usac}. The RANSAC-based algorithm has two main phases: (1) generate a hypothesis from random samples (hypothesis generation), and (2) verify it to the data (hypothesis evaluation/model verification) \cite{choi2009performance,raguram2008comparative}. Before step (1), as in the case of HT-based methods, models have to be manually defined or selected. Depending on the structure of 3D scenes, in PCS, these are usually planes, spheres, or other geometric primitives that can be represented by algebraic formulas.

In hypothesis generation, RANSAC randomly chooses $N$ sample points and estimates a set of model parameters using those sample points. For example, in PCS, if the given model is a plane, then $N$ = 3 since 3 non-collinear points determine a plane. The plane model can be represented by:

\begin{equation}aX+bY+cZ+d=0\end{equation}

where $[a,b,c,d]^T$ is the parameter set to be estimated.

In hypothesis evaluation, RANSAC chooses the most probable hypothesis from all estimated parameter sets. RANSAC uses Eq. \ref{eq:eq4} to solve the selection problem, which is regarded as an optimization problem \cite{choi2009performance}:

\begin{equation} \hat{M} = \argminA_M \{\sum_{d\in \mathscr{D}} Loss(Err(d; M)) \} \label{eq:eq4} \end{equation} 
	
where $\mathscr{D}$ is data, $Loss$ represents a loss function, and $Err$ is an error function such as geometric distance.

As an advantage of random sampling, RANSAC-based algorithms do not require complex optimization or high memory resources. Compared to HT methods, efficiency and the percentage of successful detected objects are two main advantages for RANSAC in 3D PCS \cite{tarsha2007hough}. Moreover, RANSAC algorithms have the ability to process data with a high amount of noise, even outliers \cite{schnabel2007efficient}. 
For PCS, as with HT and region growing, RANSAC is widely used in plane segmentation, such as building fa\c{c}ades \cite{bauer2005segmentation,boulaassal2007automatic,adam2018h}, building roofs \cite{chen2014methodology}, and indoor scenes \cite{li2017improved}. In some fields there is demand for the segmentation of more complex structures than planes. Schnabel et al. \cite{schnabel2007efficient} proposed an automatic RANSAC-based algorithm framework to detect basic geometric shapes in unorganized point clouds. Those shapes include not only planes, but also spheres, cylinders, cones, and tori. RANSAC-based PCS segmentation algorithms were also utilized for cylinder objects in \cite{tran2015extraction} and \cite{le2018acquiring}. 

\begin{figure}
  \centering
  \includegraphics[width=.6\linewidth]{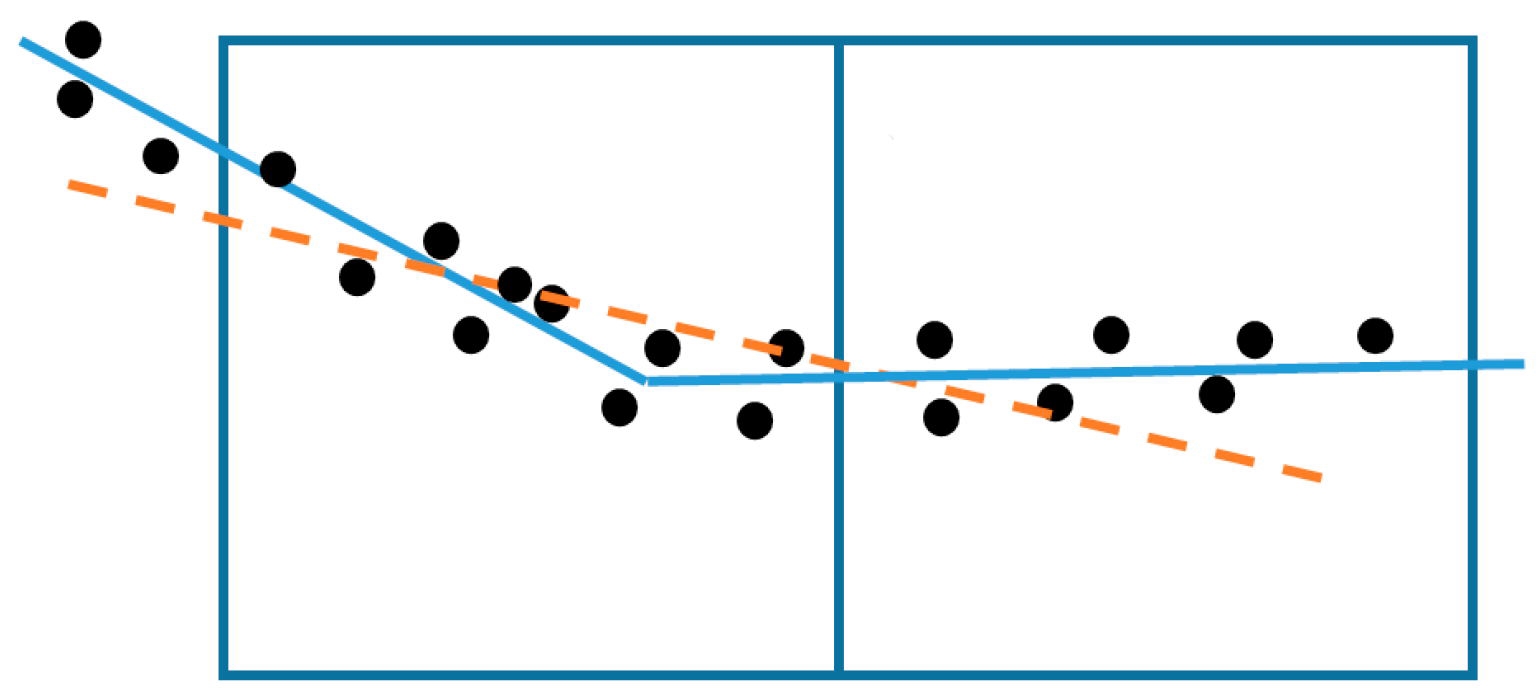}
  \caption{An example of a spurious plane \cite{li2017improved}. Two well-estimated hypothesis planes are shown in blue. A spurious plane (in orange) is generated using the same threshold.}
  \label{fig:fig1}
\end{figure}

RANSAC is a nondeterministic algorithm, and thus its main shortcoming is its spurious surface: the probability exists that models detected by RANSAC-based algorithm do not exist in reality (Fig. \ref{fig:fig1}). To overcome the adverse effect of RANSAC in PCS, a soft-threshold voting function was presented to improve the segmentation quality in \cite{xu2015investigation}, in which both the point-plane distance and the consistency between the normal vectors were taken into consideration. Li et al. \cite{li2017improved} proposed an improved RANSAC method based on NDT cells \cite{biber2003normal}, also in order to avoid spurious surface problem in 3D PCS. 

\begin{figure}
  \centering
  \includegraphics[width=.6\linewidth]{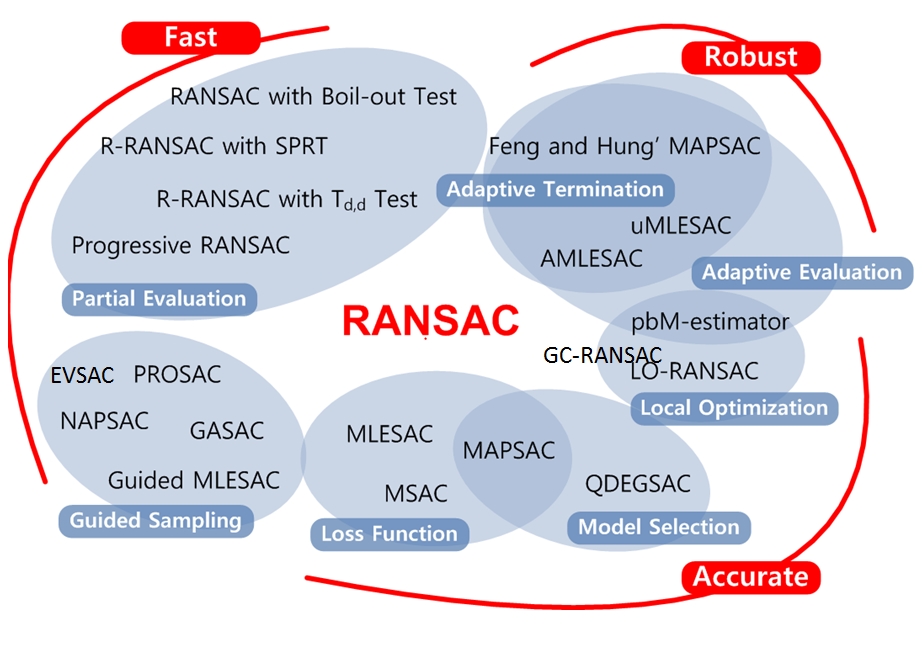}
  \caption{RANSAC family with algorithms categorized according to their performance and basic strategies \cite{choi2009performance,fragoso2013evsac,barath2018graph}. 
  }
  \label{fig:fig2}
\end{figure}

As with HT, many improved algorithms based on RANSAC have emerged over the past decades to further improve its efficiency, accuracy and robustness. These approaches have been categorized by their research objectives and are shown in Fig. \ref{fig:fig2}. The figure has been originally described in \cite{choi2009performance}, in which seven subclasses according to seven strategies are used.  Venn diagrams are utilized here to describe connections between methods and strategies, since a method may use two strategies. For detail description and explanation on those strategies, please refer to\cite{choi2009performance}. Considering that \cite{choi2009performance} is obsolete, we add two recently published methods, EVSAC \cite{fragoso2013evsac} and GC-RANSAC\cite{barath2018graph} on original figure to make it keep up with the times.

\subsection{Unsupervised clustering-based} \label{Unsupervised clustering-based}
Clustering-based methods are widely used for unsupervised PCS task. Strictly speaking, clustering-based methods are not based on a specific mathematical theory. This methodology family is a mixture of different methods that share a similar aim, which is grouping points with similar geometric features, spectral features or spatial distribution into the same homogeneous pattern. Unlike region growing and model fitting, these patterns usually are not defined in advance \cite{filin2002surface}, and thus clustering-based algorithms can be employed for irregular object segmentation, e.g., vegetation. Moreover, seed points are not required by clustering-based approaches, in contrast to region growing methods \cite{xu2018unsupervised}. 
In the early stage, $K$-means \cite{morsdorf2004lidar,sampath2006clustering,sampath2010segmentation,shahzad2012facade,zhu2014facade}, mean shift \cite{melzer2007non,ferraz20103d,shahzad2015robust,shahzad2015segmentation}, and fuzzy clustering \cite{biosca2008unsupervised,sampath2010segmentation} were the main algorithms in the clustering-based point cloud segmentation family. For each clustering approach, several similarity measures with different features can be selected, including Euclidean distance, density, and normal vector \cite{xu2018unsupervised}. From the perspective of mathematics and statistics, the clustering problem can be regarded as a graph-based optimization problem, so several graph-based methods have been experimented in PCS \cite{golovinskiy2009min,ural2012min,yan2014global}. 

\subsubsection{K-means}
$K$-means is a basic and widely used unsupervised cluster analysis algorithm. It separates the point cloud dataset into $K$ unlabeled classes. The clustering centers of $K$-means are different than the seed points of region growing. In $K$-means, every point should be compared to every cluster center in each iteration step, and the cluster centers will change when absorbing a new point. The process of $K$-means is ``clustering" rather than ``growing". It has been adopted for single tree crown segmentation on ALS data \cite{morsdorf2004lidar} and planar structure extraction from roofs \cite{sampath2006clustering}. Shahzad et al. \cite{shahzad2012facade} and Zhu et al. \cite{zhu2014facade} utilized $K$-means for building fa\c{c}ade segmentation on TomoSAR point clouds.

One advantage of $K$-means is that it can be easily adapted to all kinds of feature attributes, and can even be used in a multidimensional feature space. The main drawback of $K$-means is that it is sometimes difficult to predefine the value of $K$ properly. 

\subsubsection{Fuzzy clustering}
Fuzzy clustering algorithms are improved versions of $K$-means. $K$-means is a hard clustering method, which means the weight of a sample point to a cluster center is either 1 or 0. In contrast, fuzzy methods use soft clustering, meaning a sample point can belong to several clusters with certain nonzero weights.

In PCS, a no-initialization framework was proposed in \cite{biosca2008unsupervised}, by combining two fuzzy algorithms, Fuzzy $C$-Means (FCM) algorithm and Possibilistic $C$-Means (PCM). This framework was tested on three point clouds, including a one-scan TLS outdoor dataset with building structures. Those experiments showed that fuzzy clustering segmentation worked robustly on planer surfaces. Sampath et al. \cite{sampath2010segmentation} employed fuzzy $K$-means for segmentation and reconstruction of building roofs from an ALS point cloud. 

\subsubsection{Mean-shift}
In contrast to $K$-means, mean-shift is a classic nonparametric clustering algorithm and hence avoids the predefined $K$ problem in $K$-means\cite{comaniciu1999mean,comaniciu2002mean,cheng1995mean}. It has been applied effectively on ALS data in urban and forest terrain \cite{melzer2007non,ferraz20103d}. Mean-shift have also been adopted on TomoSAR point clouds, enabling building fa\c{c}ades and single trees to be extracted \cite{shahzad2015robust,shahzad2015segmentation}. 

As both the cluster number and the shape of each cluster are unknown, mean-shift delivers with high-probability oversegmented result \cite{yao2009object}. Hence, it is usually used as a presegmentation step before partitioning or refinement.

\subsubsection{Graph-based}
In 2D computer vision, introducing graphs to represent data units such as pixels or superpixels has proven to be an effective strategy for the segmentation task. In this case, the segmentation problem can be transformed into a graph construction and partitioning problem. Inspired by graph-based methods from 2D, some studies have applied similar strategies in PCS and achieved results in different datasets.

For instance, Golovinskiy and Funkhouser \cite{golovinskiy2009min} proposed a PCS algorithm based on min-cut \cite{boykov2006graph}, by constructing a graph using $k$-nearest neighbors. The min-cut was then successfully applied for outdoor urban object detection \cite{golovinskiy2009min}. Ural et al. \cite{ural2012min} also used min-cut to solve the energy minimization problem for ALS PCS. Each point is considered to be a node in the graph, and each node is connected to its 3D voronoi neighbors with an edge. For the roof segmentation task, Yan et al.\cite{yan2014global} used an extended $\alpha$-expansion algorithm \cite{delong2012fast} to minimize the energy function from the PCS problem. Moreover, Yao et al.\cite{yao2009object} applied a modified normalized cut (N-cut) in their hybrid PCS method.

Markov Random Field (MRF) and Conditional Random Field (CRF) are machine learning approaches to solve graph-based segmentation problems. They are usually used as supervised methods or postprocessing stages for PCSS. Major studies using CRF and supervised MRFs belong to PCSS rather than PCS. For more information about supervised approaches, please refer to section \ref{machine learning}.

\subsection{Oversegmentation, supervoxels, and presegmentation}
To reduce the calculation cost and negative effects from noise, a frequently used strategy is to oversegment a raw point cloud into small regions before applying computationally expensive  algorithms. Voxels can be regarded as the simplest oversegmentation structures. Similar to superpixels in 2D images, supervoxels are small regions of perceptually similar voxels. Since supervoxels can largely reduce the data volume of a raw point cloud with low information loss and minimal overlapping, they are usually utilized in presegmentation before executing other computationally expensive algorithms. Once oversegments like supervoxels are generated, these are fed to postprocessing PCS algorithms rather than initial points.

The most classical point cloud oversegmentation algorithm is Voxel Cloud Connectivity Segmentation (VCCS) \cite{papon2013voxel}. In this method, a point cloud is first voxelized by the octree. Then a $K$-means clustering algorithm is employed to realize supervoxel segmentation. However, since VCCS adopts fixed resolution and relies on initialization of seed points, the quality of segmentation boundaries in a non-uniform density cannot be guaranteed. To overcome this problem, Song et al. \cite{song2014boundary} proposed a two-stage supervoxel oversegmentation approach, named Boundary-Enhanced Supervoxel Segmentation (BESS). BESS preserves the shape of the object, but it also has an obvious limitation for the assumption that points are sequentially ordered in one direction. Recently, Lin et al. \cite{lin2018toward} summarized the limitations of previous studies, and formalized oversegmentation as a subset selection problem. This method adopts an adaptive resolution to preserve boundaries, a new practice in supervoxel generation. Landrieu and Boussaha \cite{landrieu2019point} presented the first supervised framework for 3D point cloud oversegmentation, achieving significant improvements compared to \cite{papon2013voxel,lin2018toward}. For PCS tasks, several studies have been based on supervoxel-based presegmentation \cite{christoph2014object,yang2015hierarchical,xu2016segmentation,xu2017voxel,xu2018unsupervised}.  

As mentioned in section \ref{Unsupervised clustering-based}, in addition to supervoxels, other methods can also be employed as presegmentation. For example, Yao et al. \cite{yao2009object} utilized mean-shift to oversegment ALS data in urban areas.

\section{point cloud semantic segmentation techniques} \label{point cloud semantic segmentation}
The procedure of PCSS is similar to clustering-based PCS. But in contrast to non-semantic PCS methods, PCSS techniques generate semantic information for every point, and are not limited to clustering. Therefore, PCSS is usually realized by supervised learning methods, including ``regular" supervised machine learning and state-of-the-art deep learning. 

\subsection{Regular supervised machine learning} \label{machine learning}
In this section, regular supervised machine learning refers to non-deep supervised learning algorithms. Comprehensive and comparative analysis on different PCSS methods based on regular supervised machine learning has been provided by previous researchers \cite{weinmann2015semantic,landrieu2017structured,niemeyer2014contextual,vosselman2017contextual}.

Paper\cite{weinmann2015contextual} pointed out that supervised machine learning applied to PCSS could be divided into two groups. One group, individual PCSS, classifies each point or each point cluster based only on its individual features, such as Maximum Likelihood classifiers based on Gaussian Mixture Models \cite{lalonde2005scale}, Support Vector Machines \cite{zhang2013svm,li2016three}, AdaBoost \cite{lodha2007aerial,wang2015multiscale}, a cascade of binary classifiers \cite{carlberg2009classifying}, Random Forests \cite{chehata2009airborne}, and Bayesian Discriminant Classifiers \cite{khoshelham2012accuracy}. The other group is statistical contextual models, such as Associative and Non-Associative Markov Networks \cite{munoz2009contextual,shapovalov2010nonassociative,najafi2014non}, Conditional Random Fields \cite{lim20093d,schmidt2012classification,niemeyer2012conditional,niemeyer2014contextual,vosselman2017contextual}, Simplified Markov Random Fields \cite{lu2012simplified}, multistage inference procedures focusing on point cloud statistics and relational information over different scales \cite{xiong20113}, and spatial inference machines modeling mid- and long-range dependencies inherent in the data \cite{shapovalov2013spatial}. 

\begin{figure}
  \centering
   \includegraphics[width=\linewidth]{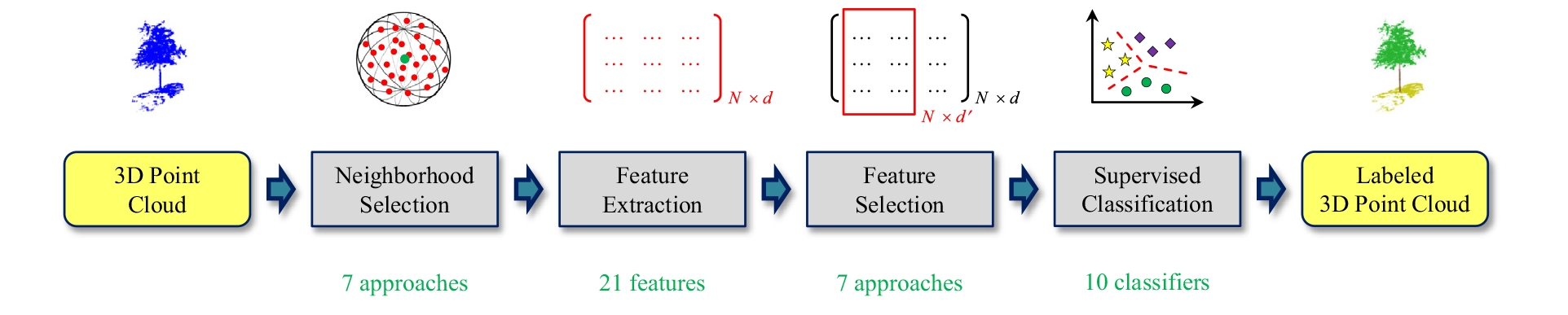}
  \caption{The PCSS framework by \cite{weinmann2015semantic}. The term ``semantic segmentation" in our review is defined as ``supervised classification" in \cite{weinmann2015semantic}.}
  \label{fig:fig3}
\end{figure}

The general procedure of the individual classification for PCSS has been well described in \cite{weinmann2015semantic}. As Fig. \ref{fig:fig3} shows, the procedure entails four stages: neighborhood selection, feature extraction, feature selection, and semantic segmentation. For each stage, paper \cite{weinmann2015semantic} summarized several crucial methods and tested different methods on two datasets to compare their performance. According to the authors' experiment, in individual PCSS, the Random Forest classifier had a good trade-off between accuracy and efficiency on two datasets. It should be noted that \cite{weinmann2015semantic} used a so-called ``deep learning" classifier in their experiments, but that is an old neural network appearing in the time of regular machine learning, not the recent deep learning methods described in section \ref{deep learning}.

\begin{figure}
  \centering
  \includegraphics[width=\linewidth]{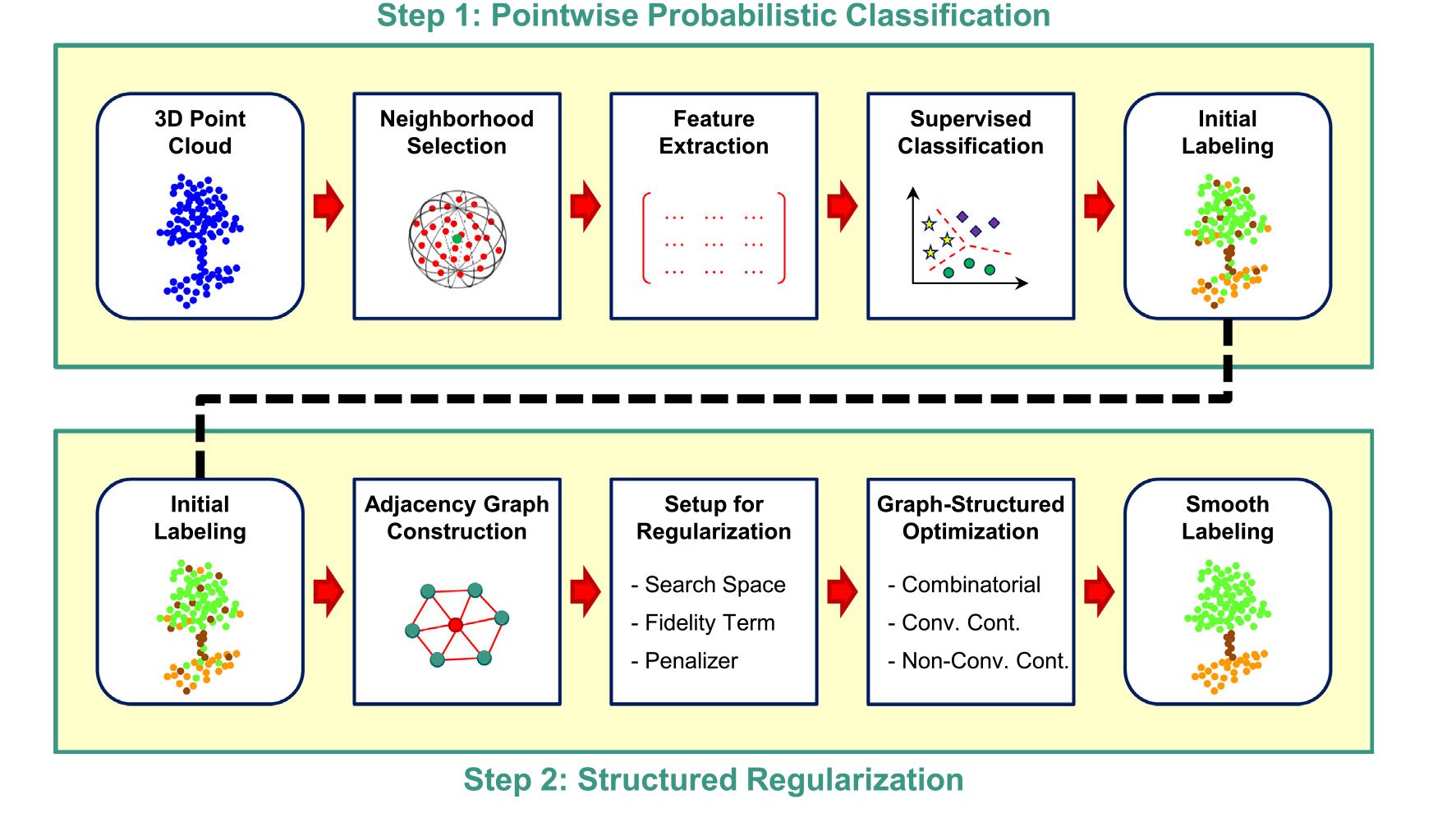}
  \caption{The PCSS framework by \cite{landrieu2017structured}. The term ``semantic segmentation" in our review is defined as ``supervised classification" in \cite{landrieu2017structured}.}
  \label{fig:fig4}
\end{figure}

Since individual PCSS does not take contextual features of points into consideration, individual classifiers work efficiently but generate unavoidable noise that cause unsmooth PCSS results. Statistical context models can mitigate this problem. Conditional Random Fields (CRF) is the most widely used context model in PCSS. Niemeyer et al. \cite{niemeyer2014contextual} provided a very clear introduction about how CRF has been used on PCSS, and tested several CRF-based approaches on the Vaihingen dataset. Based on the individual PCSS framework \cite{weinmann2015semantic}, Landrieu et al. \cite{landrieu2017structured} proposed a new PCSS framework that combines individual classification and context classification. As shown in Fig. \ref{fig:fig4}, in this framework a graph-based contextual strategy was introduced to overcome the noise problem of initial labeling, from which the process was named structured regularization or ``smoothing". 

For the regularization process, Li et al. \cite{li2016three} utilized a multilabel graph-cut algorithm to optimize the initial segmentation result from Support Vector Machine (SVM). Landrieu et al. \cite{landrieu2017structured} compared various postprocess methods in their studies, which proved that regularization indeed improved the accuracy of PCSS.

\subsection{Deep learning} \label{deep learning}
Deep learning is the most influential and fastest-growing current technique in pattern recognition, computer vision, and data analysis\cite{zhu2017deep}. As its name indicates, deep learning uses more than two hidden layers to obtain high-dimension features from training data, while traditional handcrafted features are designed with domain-specific knowledge. Before being applied in 3D data, deep learning appeared as an effective power in a variety of tasks in 2D computer vision and image processing, such as image recognition \cite{simonyan2014very,he2016deep}, object detection \cite{girshick2015fast,ren2015faster}, and semantic segmentation \cite{long2015fully,chen2018deeplab}. It has been attracting more interest in 3D analysis since 2015, driven by the multiview-based idea proposed by \cite{su2015multi}, and voxel-based 3D Convolutional Neural Network (CNN) by \cite{maturana2015voxnet}.

Standard convolutions originally designed for raster images cannot easily be directly applied to PCSS, as the point cloud is unordered and unstructured/irregular/non-raster. Thus, in order to solve this problem, a transformation of the raw point cloud becomes essential. Depending on the format of the data ingested into neural networks, deep learning-based PCSS approaches can be divided into three categories: multiview-based, voxel-based, and point-based, respectively.

\subsubsection{Multiview-based}
One of the early solutions to applying deep learning in 3D is dimensionality reduction. In short, the 3D data is represented by multi-view 2D images, which can be processed based on 2D CNNs. Subsequently, the classification results can be restored into 3D. The most influential multi-view deep learning in 3D analysis is MVCNN \cite{su2015multi}. Although the original MVCNN algorithm did not experiment on PCSS, it is a good example for learning about the multiview concept.

\begin{figure}
  \centering
  \includegraphics[width=\linewidth]{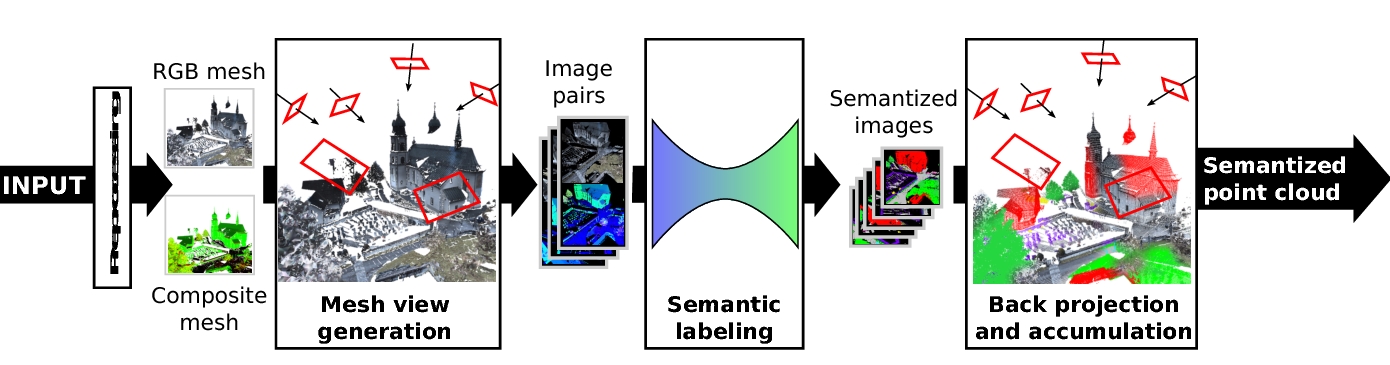}
  \caption{The Workflow of SnapNet \cite{boulch2018snapnet}.}
  \label{fig:fig5}
\end{figure}

The multiview-based methods have solved the structuring problems of point cloud data well, but there are two serious shortcomings in these methods. Firstly, they cause numerous limitations and a loss in geometric structures, as 2D multiview images are just an approximation of 3D scenes. As a result, complex tasks such as PCSS could yield limited and unsatisfactory performances. Secondly, multiview projected images must cover all spaces containing points. For large, complex scenes, it is difficult to choose enough proper viewpoints for multiview projection. Thus, few studies used multiview-based deep learning architecture for PCSS. One of exceptions is SnapNet \cite{boulch2017unstructured,boulch2018snapnet}, which uses full dataset semantic-8 of semantic3D.net as the test dataset. Fig. \ref{fig:fig5} shows the workflow of SnapNet. In SnapNet, the preprocessing step aims at decimating the point cloud, computing point features and generating a mesh. Snap generation is to generate RGB images and depth composite images of the mesh, based on various virtual cameras. Semantic labeling is to realize image semantic segmentation from the two types of input images, by image-based deep learning. The last step is to project 2D semantic segmentation results back to 3D space, thereby 3D semantics can be acquired.

\subsubsection{Voxel-based}
Combining voxels with 3D CNNs is the other early approach in deep learning-based PCSS. Voxelization solves both unordered and unstructured problems of the raw point cloud. Voxelized data can be further processed by 3D convolutions, as in the case of pixels in 2D neural networks. 

Voxel-based architectures still have serious shortcomings. In comparison to the point cloud, the voxel structure is a low-resolution form. Obviously, there is a loss in data representation. In addition, voxel structures not only store occupied spaces, but also store free or unknown spaces, which can result in high computational and memory requirements. 

\begin{figure}
  \centering
  \includegraphics[width=\linewidth]{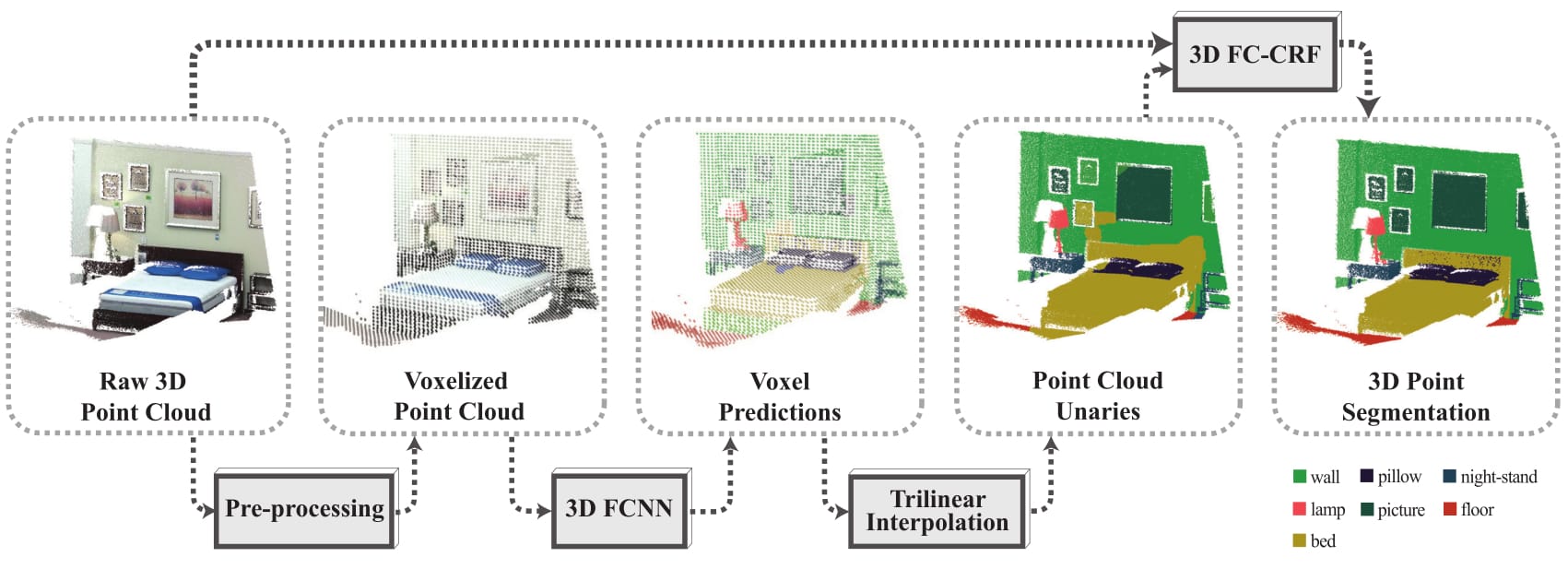}
  \caption{The Workflow of SegCloud \cite{tchapmi2017segcloud}.}
  \label{fig:fig6}
\end{figure}

The most well-known voxel-based 3D CNN is VoxNet \cite{maturana2015voxnet}, but this was only tested for object detection. On the PCSS task, some papers, like  \cite{riegler2017octnet,wang2017cnn,tchapmi2017segcloud} and \cite{meng2018vv}, proposed representative frameworks. SegCloud \cite{tchapmi2017segcloud} is an end-to-end PCSS framework that combines 3D-FCNN, trilinear interpolation (TI), and fully connected Conditional Random Fields (FC-CRF) to accomplish the PCSS task. Fig. \ref{fig:fig6} shows the framework of SegCloud, which also provides a basic pipeline of voxel-based semantic segmentation. In SegCloud, the preprocessing step is to voxelize raw point clouds. Then a 3D fully convolutional neural netwotk is applied to generate downsampled voxel labels. After that, a trilinear interpolation layer is employed to transfer voxel labels back to 3D point labels. Finally, a 3D fully connected CRF method is utilized to regularize previous 3D PCSS results, and acquire final results. SegCloud used to be the state-of-art approach in both S3DIS and semantic3D.net, but it did not take any steps to optimize high computational and memory problem from fixed-sized voxels. With more advanced methods springing up, SegCloud has fallen from favor in recent years.

To reduce unnecessary computation and memory consumption, the flexible octree structure is an effective replacement for fixed-size voxels in 3D CNNs. OctNet \cite{riegler2017octnet} and O-CNN \cite{wang2017cnn} are two representative approaches. Recently, VV-NET \cite{meng2018vv} extended the use of voxels. VV-Net utilized a radial basis function-based Variational Auto-Encoder (VAE) network, which provided a more information-rich representation for point cloud compared with binary voxels. What is more, Choy et al. \cite{choy20194d} proposed 4-dimensional convolutional neural networks (MinkowskiNets) to process 3D-videos,  which are a series of CNNs for high-dimensional spaces including the 4D spatio-temporal data. MinkowskiNets can also be applied on 3D PCSS tasks. They have achieved good performance on a series of PCSS benchmark datasets, especially a significant accuracy improvement on ScanNet \cite{dai2017scannet}.

\subsubsection{Directly process point cloud data}
As there are serious limitations in both multiview- and voxel-based methods (e.g., loss in structure resolution), exploring PCSS methods directly on point is a natural choice. Up to now, many approaches have emerged and are still emerging \cite{qi2017pointnet,qi2017pointnet++,wang2018dynamic,landrieu2018large,li2018pointcnn}. Unlike employing separated pretransformation operation in multiview-based and voxel-based cases, in these approaches the canonicalization is binding with the neural network architecture. 

\begin{figure}[!htb]
  \centering
  \includegraphics[width=\linewidth]{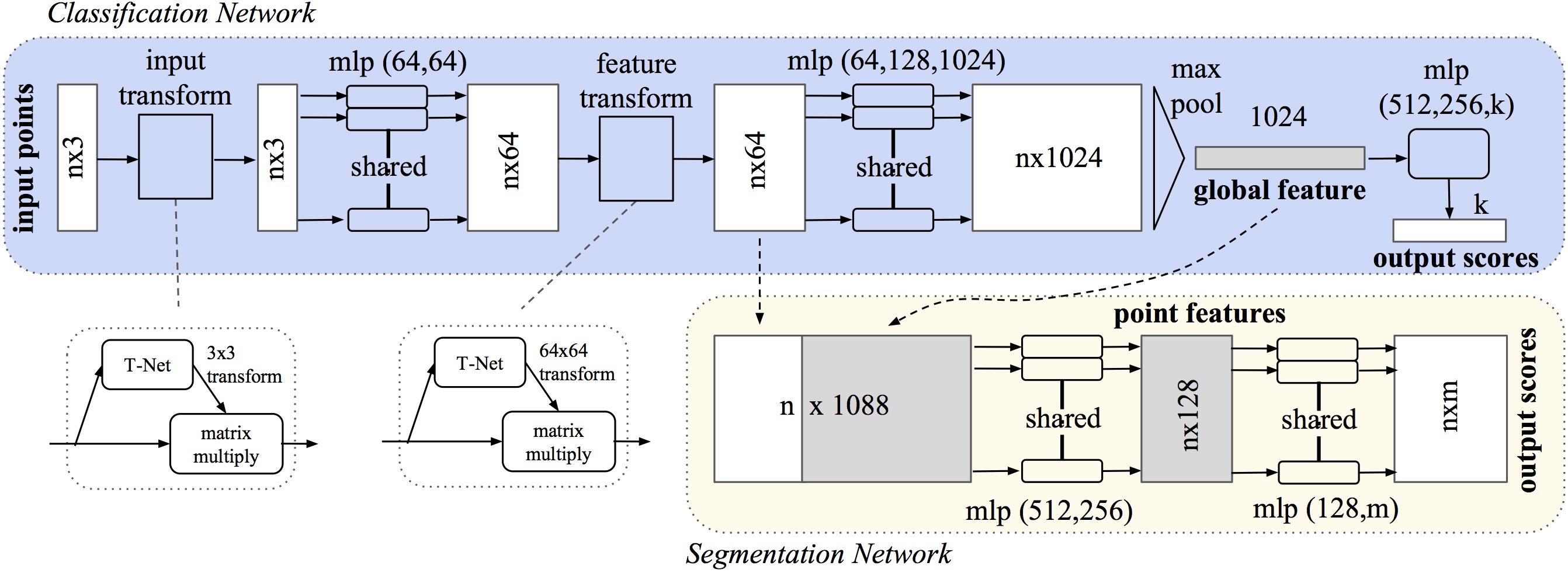}
  \caption{The Workflow of PointNet \cite{qi2017pointnet}. In this figure, ``Classification Network'' is used for object classification. ``Segmentation Network" is applied for the PCSS mission.}
  \label{fig:fig7}
\end{figure}

PointNet \cite{qi2017pointnet} is a pioneering deep learning framework which has been performed directly on point. Different with recently published point cloud networks, there is no convolution operator in PointNet. The basic principle of PointNet is: 
\begin{equation}
    f(\{x_1,...,x_n\})\approx g(h(x_1),...,h(x_n))
\end{equation}

where $f : 2^{\mathbb{R}^N}\rightarrow \mathbb{R}$
and  $h:\mathbb{R}^N\rightarrow \mathbb{R}^K$. $g:\underbrace{\mathbb{R}^K\times...\times \mathbb{R}^K}_{n} \rightarrow \mathbb{R}$ is a symmetric function, used to solve the ordering problem of point clouds.
As Fig. \ref{fig:fig7} shows, PointNet uses MultiLayer Perceptrons (MLPs) to approximate $h$, which represents the per-point local features corresponding to each point. The global features of point sets $g$ are aggregated by all per-point local features in a set, through a symmetric function, max pooling. For the classification task, output scores for $k$ classes can be produced by a MLP operation on global features. For the PCSS task, in addition to global features, per-point local features are demanded. PointNet concatenates aggregated global features and per-point local features into combined point features. Subsequently, new per-point features are extracted from the combined point features by MLPs. On their basis, semantic labels are predicted.

Although more and more newly published networks outperform PointNet on various benchmark datasets, PointNet is still a baseline for PCSS research. The original PointNet uses no local structure information within neighboring points. In a further study, Qi et al. \cite{qi2017pointnet++} used a hierarchical neural network to capture local geometric features to improve the basic PointNet model and proposed PointNet++. Drawing inspiration from PointNet/PointNet++, studies on 3D deep learning focus on feature augmentation, especially to local features/relationships among points, utilizing knowledge from other fields to improve the performance of the basic PointNet/PointNet++ algorithms. For example, Engelmann et al. \cite{engelmann2017exploring} employed two extensions on the PointNet to incorporate larger-scale spatial context. Wang et al. \cite{wang2018dynamic} considered that missing local features was still a problem in PointNet++, since it neglected the geometric relationships between a single point and its neighbors. To overcome this problem, Wang et al. \cite{wang2018dynamic} proposed Dynamic Graph CNN (DGCNN). In this network, the authors designed a procedure called EdgeConv to extract edge features while maintaining permutation invariance. Inspired by the idea of the attention mechanism, Wang et al. \cite{wang2019graph} designed a Graph Attention Convolution (GAC), of which kernels could be dynamically adapted to the structure of an object. GAC can capture the structural features of point clouds while avoiding feature contamination between objects. To exploit richer edge features, Landrieu and Simonovsky \cite{landrieu2018large} introduced the SuperPoint Graph (SPG), offering both compact and rich representation of contextual relationships among object parts rather than points. The partition of the superpoint can be regarded as a nonsemantic presegmentation step. After SPG construction, each superpoint is embedded in a basic PointNet network and then refined in Gated Recurrent Units (GRUs) for PCSS. Benefiting from information-rich  downsampling, SPG is highly efficient for large-volume datasets. 

 Also in order to overcome the drawback of no local features represented by neighboring points in PointNet, 3P-RNN \cite{ye20183d} adopted a Pointwise Pyramid Pooling (3P) module to capture the local feature of each point. In addition, it employed a two-direction Recurrent Neural Network (RNN) model to integrate long-range context in PCSS tasks. The 3P-RNN technique has increased overall accuracy at a negligible extra overhead. Komarichev et al. \cite{komarichev2019cnn} introduced an annular convolution, which could capture the local neighborhood by specifying the ring-shaped structures and directions in the computation, and adapt to the geometric variabil1ity and scalability at the signal processing level. Due to the fact that the $K$-nearest neighbor search in PointNet++ may lead to the $K$ neighbors falling in one orientation, Jiang et al. \cite{jiang2018pointsift} designed PointSIFT to capture local features from eight orientations. In the whole architecture, the PointSIFT module achieves multiscale representation by stacking several Orientation-Encoding (OE) units. The PointSIFT module can be integrated into all kinds of PointNet-based 3D deep learning architectures to improve the representational ability for 3D shapes. Built upon PointNet++, PointWeb \cite{zhao2019pointweb} utilized the Adaptive Feature Adjustment (AFA) module to find the interaction between points. The aim of AFA is also to capture and aggregate local features of points.

Besides, based on PointNet/PointNet++, instance segmentation can also be realized, even accompanied by PCSS. For instance, Wang et al. \cite{wang2018sgpn} presented the Similarity Group Proposal Network (SGPN). SGPN is the first published point cloud instance segmentation framework. Yi et al. \cite{yi2019gspn} presented a Region-based PointNet (R-PointNet). The core module of R-PointNet is named as Generative Shape Proposal Network (GSPN), of which the base is PointNet. Pham et al. \cite{pham2019jsis3d} applied a Multi-task Pointwise Network (MT-PNet) and a Multi-Value Conditional Random Field (MV-CRF) to address PCSS and instance segmentation simultaneously. MV-CRF jointly realized the optimization of semantics and instances. Wang et al. \cite{Wang2019Associatively} proposed an Associatively Segmenting Instances and Semantics (ASIS) module, making PCSS and instance segmentation take advantage of each other, leading to a win-win situation. In \cite{Wang2019Associatively}, the backbone that networks employed are also PointNet and PointNet++.

An increasing number of researchers have chosen an alternative to PointNet, employing the convolution as a fundamental and significant component. Some of them, like \cite{wang2018dynamic,wang2019graph,komarichev2019cnn}, have been introduced above. In addition, PointCNN used a $\mathcal{X}$-transformation instead of symmetric functions to canonicalize the order \cite{li2018pointcnn}, which is a generalization of CNNs to feature learning from unorderd and unstructured point clouds. Su et al. \cite{su2018splatnet} provided a PCSS framework that could fuse 2D images with 3D point clouds, named SParse LATtice Networks (SPLATNet), preserving spatial information even in sparse regions. Recurrent Slice Networks (RSN) \cite{huang2018recurrent} exploited a sequence of multiple 1$\times$1 convolution layers for feature learning, and a slice pooling layer to solve the unordered problem of raw point clouds. A RNN model was then applied on ordered sequences for the local dependency modeling. Te et al. \cite{te2018rgcnn} proposed Regularized Graph CNN (RGCNN) and tested it on a part segmentation dataset, ShapeNet \cite{chang2015shapenet}. Experiments show that RGCNN can reduce computational complexity and is robust to low density and noise. Regarding convolution kernels as nonlinear functions of the local coordinates of 3D points comprised of weight and density functions, Wu et al. \cite{Wu2019PointConv} presented PointConv. PointConv is an extension to the Monte Carlo approximation of the 3D continuous convolution operator. PCSS is realized by a deconvolution version of PointConv.

As SPG \cite{landrieu2018large}, DGCNN \cite{wang2018dynamic}, RGCNN \cite{te2018rgcnn} and GAC \cite{wang2019graph} employed graph structures in neural networks, they can also be regarded as Graph Neural Networks (GNNs) in 3D \cite{zhou2018graph,wu2019comprehensive}. 

The research on PCSS based on deep learning is still ongoing. New ideas and approaches on the topic of 3D deep learning-based frameworks are keeping popping up. Current achievements have proved that it is a great boost for the accuracy of 3D PCSS.

\subsection{Hybrid methods} \label{hybrid methods}
In PCSS, hybrid segment-wise methods have been attracting researchers' attention in recent years. A hybrid approach is usually made up of at least two stages: (1) utilize an oversegmentation or PCS algorithm (introduced in section \ref{point cloud segmentation techniques}) as the presegmentation, and (2) apply PCSS on segments from (1) rather than points. In general, as with presegmentation in PCS, presegmentation in PCSS also has two main functions: to reduce the data volume and to conduct local features. Oversegmentation for supervoxels is a kind of presegmentation algorithm in PCSS \cite{lim20093d}, since it is an effective way to reduce the data volume with light accuracy loss. In addition, because nonsemantic PCS methods can provide rich natural local features, some PCSS studies also use them as presegmentation. For example, Zhang et al. \cite{zhang2013svm} employed region growing before SVM. Vosselman et al. \cite{vosselman2017contextual} applied HT to generate planar patches in their PCSS algorithm framework as the presegmentation. In deep learning, Landrieu and Simonovsky \cite{landrieu2018large} exploited a superpoint structure as the presegmentation step, and provided a contextual PCSS network combining superpoint graphs with PointNet and contextual segmentation. Landrieu and Boussaha \cite{landrieu2019point} used a supervised algorithm to realize the presegmentation, which is the first supervised framework for 3D point cloud oversegmentation.

\section{Discussion} \label{Discussion}

\subsection{Open issues in segmentation techniques}

\subsubsection{Features}
One of the core questions in pattern recognition is how to obtain effective features. Essentially, the biggest differences among the various methods in PCSS or PCS are the differences of feature design, selection, and application. Feature selection is a trade-off between algorithm accuracy and efficiency. Focusing on PCSS, Weinmann et al. \cite{weinmann2015contextual} analyzed features from three aspects: neighborhood selection (fixed or individual); feature extraction (single-scale or multi-scale); and classifier selection (individual classifier or contextual classifier). Deep learning-based algorithms face similar problems. The local feature is a significant aspect to be improved after the birth of PointNet \cite{qi2017pointnet}. 

Even in a PCS task, different methods also show different understandings of features. Model fitting is actually searching for a group of points connected with certain geometric primitives, which also can be defined as features. For this reason, deep learning has been introduced into model fitting recently \cite{li2019supervised}. The criteria or the similarity measure in region growing or clustering is the feature of a point essentially. The improvement of an algorithm reflects its ability to more strongly capture features.

\subsubsection{Hybrid}
As mentioned in section \ref{hybrid methods}, hybrid is a strategy for PCSS. Presegmentation can provide local features in a natural way. Once the development of neural network architectures stabilizes, nonsemantic presegmentation might become a progressive course for PCSS. 

\subsubsection{Contextual information}
In PCSS tasks, contextual models are crucial tools for regular supervised machine learning, widely exploited as a smoothing postprocessing step. In deep learning, several methods, like \cite{tchapmi2017segcloud}, \cite{landrieu2018large}, \cite{pham2019jsis3d} and \cite{choy20194d}, have employed contextual segmentation, but there is still room for further improvements. 

\subsubsection{PCSS with GNNs}
GNN is becoming increasingly popular in 2D image processing \cite{zhou2018graph,wu2019comprehensive}. For PCSS tasks, its excellent performance has been shown in \cite{landrieu2018large}, \cite{wang2018dynamic}, \cite{te2018rgcnn} and \cite{wang2019graph}. Similar to contextual models, the GNN might also have some surprises for PCSS. But more research is required in order to evaluate its performance.

\subsubsection{Regular machine learning vs. deep learning}
Before deep learning emerged, regular machine learning was the choice of supervised PCSS. Deep learning has changed the way a point cloud is handled. Compared with regular machine learning, deep learning has notable advantages: (1) it is more efficient at handling large-volume datasets; (2) there is no need to handcraft feature design and selection, a difficult task in regular machine learning; and (3) it yields high ranks (high-accuracy results) on public benchmark datasets. Nevertheless, deep learning is not a universal solution. Firstly, its principal shortcoming is poor interpretability. Currently, it is well known how each type of layers (e.g., convolution, pooling) works in a neural network. In pioneering PCSS works, such knowledge has been used to develop a series of functional networks \cite{qi2017pointnet,li2018pointcnn,Wu2019PointConv}. However, a detailed internal decision-making process for deep learning is not yet understood, and therefore cannot be fully described. As a result, certain fields demanding high-level safety or stability cannot trust deep learning completely. A typical example that is relevant to PCSS is autonomous driving. Secondly, data limit the application of deep learning-based PCSS. Compared with annotating 2D images, acquiring and annotating a point cloud is much more complicated. Finally, although current public datasets provide several indoor and outdoor scenes, they cannot meet the demand in real applications sufficiently. 

\subsection{Remote sensing meets computer vision}
Remote sensing and general computer vision might be two of the most active groups interested in point clouds, having published many pioneering studies. The main difference between these two groups is that computer vision focuses on new algorithms to further improve the accuracy of the results. Remote sensing researchers, on the other hand, are trying to apply these techniques on different types of datasets. However, in many cases the algorithms proposed by computer vision studies cannot be adopted in remote sensing directly. 

\subsubsection{Evaluation system}
In generic computer vision, in order to evaluate the accuracy, the overall accuracy is a significant index. However, some remote sensing applications care more about the accuracy of certain objects. For instance, for urban monitoring the accuracy of buildings is crucial, while the segmentation or the semantic segmentation of other objects is less important. Thus, compared to computer vision, remote sensing needs a different evaluation system for selecting proper algorithms.

\subsubsection{Multi-source Data}
As discussed in section \ref{An Introduction to Point Cloud}, point clouds in remote sensing and computer vision appear differently. For example, airborne/spaceborne 2.5D and/or sparse point clouds are also crucial components of remote sensing data, while computer vision focuses on denser full 3D.

\subsubsection{Remote sensing algorithms}
Published computer vision algorithms are usually tested on a small-area dataset with limited categories of objects. However, for remote sensing applications, large-area data with more complex and specific ground object categories are demanded. For example, in agricultural remote sensing, vegetation is expected to be separated into certain specific species, which is difficult for current computer vision algorithms to solve.

\subsubsection{Noise and outliers}
Current computer vision algorithms do not pay much attention to noise, while in remote sensing, sensor noise is unavoidable. Currently, noise adaptive algorithms are unavailable. 

\subsection{Limitation of public benchmark datasets}
In section \ref{Benchmark datasets}, several popular benchmark datasets are listed. Obviously, in comparison to the situation several years ago, the number of large-scale datasets with dense point clouds and rich information available to researchers has increased considerably. Some datasets, such as semantic3D.net and S3DIS, have hundreds of millions of points. However, those benchmark datasets are still insufficient for PCSS tasks. 

\subsubsection{Limited data types}
Despite the fact that several large datasets for PCSS are available, there is still demand for more varied data. In the real world, there are much more object categories than the ones considered in current benchmark datasets. For example, semantic3D.net provides a large-scale urban point cloud benchmark. However, it only covers one kind of cities. If researchers chose a different city for a PCSS task, in which building styles, vegetation species, and even ground object types would differ, algorithm results might in turn be different.

\subsubsection{Limited data sources}
Most mainstream point cloud benchmark datasets are acquired from either LiDAR or RGB-D. But in practical applications, image-derived point clouds cannot be ignored. As previously mentioned, in remote sensing the airborne 2.5D point cloud is an important category, but for PCSS tasks only the Vaihingen dataset \cite{rottensteiner2013isprs,niemeyer2014contextual} is published as a benchmark dataset. New data types, such as satellite photogrammetric point clouds, InSAR point clouds, and even multi-source fusion data, are also necessary to establish corresponding baselines and standards. 

\section{Conclusion} \label{Conclusion}
This paper provided a review of current PCSS and PCS techniques. This review not only summarizes the main categories of relevant algorithms, but also briefly introduces the acquisition methodology and evolution of point clouds. In addition, the advanced deep learning methods that have been proposed in recent years are compared and discussed. Due to the complexity of the point cloud, PCSS is more challenging than 2D semantic segmentation. Although many approaches are available, they have each been tested on very limited and dissimilar datasets, so it is difficult to select the optimal approach for practical applications. Deep learning-based methods have ranked high for most of the benchmark-based evaluations, yet there is no standard neural network publicly available. Improved neural networks for the solution of PCSS problems can be expected to be designed in coming years.

Most current methods have only considered point features, but in practical applications such as remote sensing the noise and outliers are still problems that cannot be avoided. Improving the robustness of current approaches, and combining initial point-based algorithms with different sensor theories to denoise the data are two potential future fields of research for semantic segmentation.

\section*{Acknowledgment}
The authors would like to thank Dr. D. Cerra and P. Schwind for proof-reading this paper, and the anonymous reviewers and the associate editor for commenting and improving this paper.

The work of Yuxing Xie is supported by the DLR-DAAD research fellowship (No. 57424731), which is funded by the German Academic Exchange Service (DAAD) and the German Aerospace Center (DLR).

The work of Xiao Xiang Zhu is jointly supported by the European Research Council (ERC) under the European Union's Horizon 2020 research and innovation programme (grant agreement No. [ERC-2016-StG-714087], Acronym: \textit{So2Sat}), Helmholtz Association under the framework of the Young Investigators Group ``SiPEO'' (VH-NG-1018, www.sipeo.bgu.tum.de), and the Bavarian Academy of Sciences and Humanities in the framework of Junges Kolleg.

\ifCLASSOPTIONcaptionsoff
  \newpage
\fi



%


\bibliography{References} 
\bibliographystyle{ieeetr}


%

\begin{IEEEbiographynophoto}{Yuxing Xie}
(yuxing.xie@dlr.de) received the B.Eng. degree in remote sensing science and technology from Wuhan University, Wuhan, China, in 2015, and the M.Eng. degree in photogrammetry and remote sensing from Wuhan University, Wuhan, China, in 2018. He is currently pursuing the Ph.D. degree with the Remote Sensing Technology Institute, German Aerospace Center (DLR), We{\ss}ling, Germany, and the Technical University of Munich (TUM), Munich, Germany. His research interests include point cloud processing and the application of 3D geographic data.
\end{IEEEbiographynophoto}
\begin{IEEEbiographynophoto}{Jiaojiao Tian}
(jiaojiao.tian@dlr.de) received her B.S in Geo-Information Systems at the China University of Geoscience (Beijing) in 2006, M. Eng in Cartography and Geo-information at the Chinese Academy of Surveying and Mapping (CASM) in 2009, and Ph.D. degree in mathematics and computer science from Osnabr{\"u}ck University, Germany in 2013. Since 2009, she has been with the Photogrammetry and Image Analysis Department, Remote Sensing Technology Institute, German Aerospace Center, We{\ss}ling, Germany, where she is currently the Head of the 3D Modeling Team. In 2011, she was a Guest Scientist with the Institute of Photogrammetry and Remote Sensing, ETH Zürich, Zurich, Switzerland. Her research interests include 3-D change detection, digital surface model (DSM) generation, 3D point cloud semantic segmentation, object extraction, and DSM-assisted building reconstruction and classification.
\end{IEEEbiographynophoto}
\begin{IEEEbiographynophoto}{Xiao Xiang Zhu}
(xiaoxiang.zhu@dlr.de) received the Master (M.Sc.) degree, her doctor of engineering (Dr.-Ing.) degree and her “Habilitation” in the field of signal processing from Technical University of Munich (TUM), Munich, Germany, in 2008, 2011 and 2013, respectively.
\par
  She is currently the Professor for Signal Processing in Earth Observation (www.sipeo.bgu.tum.de) at Technical University of Munich (TUM) and German Aerospace Center (DLR); the head of the department ``EO Data Science'' at DLR's Earth Observation Center; and the head of the Helmholtz Young Investigator Group ``SiPEO'' at DLR and TUM. Since 2019, she is co-coordinating the Munich Data Science Research School (www.mu-ds.de). She is also leading the Helmholtz Artificial Intelligence Cooperation Unit (HAICU) -- Research Field ``Aeronautics, Space and Transport". Prof. Zhu was a guest scientist or visiting professor at the Italian National Research Council (CNR-IREA), Naples, Italy, Fudan University, Shanghai, China, the University  of Tokyo, Tokyo, Japan and University of California, Los Angeles, United States in 2009, 2014, 2015 and 2016, respectively. Her main research interests are
  remote sensing and Earth observation, signal processing, machine learning and data science, with a special application focus on global urban mapping. 

  Dr. Zhu is a member of young academy (Junge Akademie/Junges Kolleg) at the Berlin-Brandenburg Academy of Sciences and Humanities and the German National  Academy of Sciences Leopoldina and the Bavarian Academy of Sciences and Humanities. She is an associate Editor of IEEE Transactions on Geoscience and Remote Sensing.
\end{IEEEbiographynophoto}




\end{document}